%% file: main.tex
\documentclass[sigconf]{acmart}
\AtBeginDocument{%
  }

\copyrightyear{2026}
\acmYear{2026}
\setcopyright{cc}
\setcctype{by}
\acmConference[WWW '26]{Proceedings of the ACM Web Conference 2026}{April 13--17, 2026}{Dubai, United Arab Emirates}
\acmBooktitle{Proceedings of the ACM Web Conference 2026 (WWW '26), April 13--17, 2026, Dubai, United Arab Emirates}
\acmPrice{}
\acmDOI{10.1145/3774904.3792579}
\acmISBN{979-8-4007-2307-0/2026/04}

\usepackage{enumitem}
\usepackage{amsmath}
\usepackage{makecell}
\usepackage{algorithm}
\usepackage{algpseudocode}
\usepackage{subfigure}
\usepackage[table]{xcolor}
\usepackage{multirow}
\usepackage{hyperref}
\usepackage{balance}

\newcommand{\method}{\textsf{FedRecGEL}}
\newtheorem{lemma}{lemma}

\begin{document}

\title{Sharpness-Aware Minimization for Generalized Embedding Learning in Federated Recommendation}

\author{Fengyuan Yu}
\affiliation{%
  \institution{Zhejiang University}
  \city{Hangzhou}
  \country{China}}
\email{fengyuanyu@zju.edu.cn}

\author{Xiaohua Feng}
\affiliation{%
  \institution{Zhejiang University}
  \city{Hangzhou}
  \country{China}}
\email{fengxiaohua@zju.edu.cn}

\author{Yuyuan Li}
\affiliation{%
  \institution{Hangzhou Dianzi University}
  \city{Hangzhou}
  \country{China}}
\email{y2li@hdu.edu.cn}

\author{Changwang Zhang}
\affiliation{%
  \institution{OPPO Research Institute}
  \city{Shenzhen}
  \country{China}}
\email{changwangzhang@foxmail.com}

\author{Jun Wang}
\authornote{Corresponding authors (in order): Chaochao Chen and Jun Wang.}
\affiliation{%
  \institution{OPPO Research Institute}
  \city{Shenzhen}
  \country{China}}
\email{junwang.lu@gmail.com}

\author{Chaochao Chen}
\authornotemark[1]
\affiliation{%
  \institution{Zhejiang University}
  \city{Hangzhou}
  \country{China}}
\email{zjuccc@zju.edu.cn}

\renewcommand{\shortauthors}{Fengyuan Yu et al.}

\begin{abstract}
Federated recommender systems enable collaborative model training while keeping user interaction data local and sharing only essential model parameters, thereby mitigating privacy risks.
However, existing methods overlook a \textbf{critical issue}, i.e., the stable learning of a generalized item embedding throughout the federated recommender system training process.
Item embedding plays a central role in facilitating knowledge sharing across clients.
Yet, under the cross-device setting, local data distributions exhibit significant heterogeneity and sparsity, exacerbating the difficulty of learning generalized embeddings.
These factors make the stable learning of generalized item embeddings both indispensable for effective federated recommendation and inherently difficult to achieve.
To fill this gap, we propose a new federated recommendation framework, named \textbf{Fed}erated \textbf{Rec}ommendation with \textbf{G}eneralized \textbf{E}mbedding \textbf{L}earning (\method{}).
We reformulate the federated recommendation problem from an item-centered perspective and cast it as a multi-task learning problem, aiming to learn generalized embeddings throughout the training procedure.
Based on theoretical analysis, we employ sharpness-aware minimization to address the generalization problem, thereby stabilizing the training process and enhancing recommendation performance.
Extensive experiments on four datasets demonstrate the effectiveness of \method{} in significantly improving federated recommendation performance.
Our code is available at \url{https://github.com/anonymifish/FedRecGEL}.
\end{abstract}




\maketitle

\input{body/01_introduction}

\input{body/02_related_works}

\input{body/03_preliminary}

\input{body/04_method}

\input{body/05_experiment}

\input{body/06_conclusion}

\begin{acks}
This work was supported in part by the National Key R\&D Program of China (2022YFB4501500,2022YFB4501504), the National Natural Science Foundation of China (No.~62522217), and the National Natural Science Foundation of China (No.~62402148).
\end{acks}

\bibliographystyle{ACM-Reference-Format}
\balance
\bibliography{reference}

\appendix

\input{body/appendix}

\end{document}

%% file: body/01_introduction.tex
\section{Introduction}
Recommender systems~\cite{isinkaye2015recommendation, zhang2019deep, ko2022survey, wang2025drvae, qian2025personalized, yang2024federated, zheng2023decentralized} are designed to leverage user-item interaction information to provide personalized content to users.
They play a pivotal role across diverse domains, helping users discover items that match their preferences, thus improving user experience and driving economic value.
Conventional recommender systems typically rely on centralized collection of user behavior data, which raises significant privacy concerns~\cite{voigt2017eu}.
To mitigate these issues, federated learning~\cite{shen2025build,shen2025mars} is introduced, and federated recommender systems~\cite{mcmahan2017communication, sun2024survey} have emerged as a promising solution.
In such a system, each client holds user interaction data locally and shares only essential model parameters for collaborative training.
Federated recommender systems operating under the cross-device setting~\cite{chen2023fs, zhou2023fastpillars} involve a vast number of clients, where each client corresponds to an individual user.
This cross-device setting further exacerbates data heterogeneity and sparsity, thereby substantially increasing the difficulty of training.

Existing work on federated recommendation generally falls into two categories aimed at improving the training process and thus enhancing recommendation performance.
\emph{First}, a number of studies focus on clustering similar users based on user embeddings or item embeddings updated by clients, and then conducting collaborative training within each cluster~\cite{he2024co, luo2022personalized, mao2024cluster, muhammad2020fedfast, luo2024perfedrec++}.
These approaches seek to stabilize the training process by grouping clients with relatively small data distributional differences, thereby alleviating the effects of data heterogeneity and sparsity within clusters.
\emph{Second}, other studies aim to train personalized models for each client by adapting the global model through local fine-tuning~\cite{zhang2023dual, zhang2024gpfedrec, lin2025cec} or by fusing the global and local models~\cite{li2023federated, li2025personalized, yao2024swift, zhou2025joint}.
These methods allocate more model parameters to capture client-specific data distribution, mitigating the negative impact of heterogeneity across clients.

However, both of the aforementioned categories of methods still overlook a \textbf{critical challenge}, i.e., the stable learning of a generalized item embedding throughout the federated recommender system training process.
In federated recommender systems, the same item can be interacted with by multiple users, and the corresponding updates are aggregated on the server to capture global information, which is then distributed back to the clients for local training, thereby facilitating knowledge sharing across clients.
Consequently, it is essential to learn a generalized item embedding that not only characterizes the global distribution but also adapts to diverse local distributions.
Additionally, in the cross-device setting, the local data distribution exhibits significant heterogeneity and sparsity, with only a small subset of items participating in local training, further exacerbating the difficulty of learning generalized embeddings~\cite{jiang2025tutorial, tong2025dapt, sun2019multinational}.
Taken together, these factors make the stable learning of generalized item embeddings both indispensable for effective federated recommendation and inherently difficult to achieve.
Clustering-based approaches only learn item embeddings specific to a group of similar users, while personalized methods do not focus on improving the generalization of global embeddings.

To fill this gap, we propose a new federated recommendation framework, named \textbf{Fed}erated \textbf{Rec}ommendation with \textbf{G}eneralized \textbf{E}mbedding \textbf{L}earning (\method{}).
Different from the aforementioned approaches, \method{} directly addresses the challenge of learning generalized embeddings.
\method{} adopts an item-centered perspective to formalize the federated recommendation problem as a multi-task learning problem, with the objective of improving the generalization capability of the learned embeddings.
Through theoretical analysis, we demonstrate that this generalization challenge can be effectively addressed using sharpness-aware minimization.
Building on this insight, \method{} modifies both the local training and global aggregation processes, employing sharpness-aware minimization to learn robust and generalized embeddings.
This design stabilizes the training procedure and ultimately enhances recommendation performance.

We summarize the main contributions of this paper as follows:
\begin{itemize}[leftmargin=*]\setlength{\itemsep}{-\itemsep}
    \item We reformulate the federated recommendation problem from an item-centered perspective and cast it as a multi-task learning problem, highlighting the importance of learning generalized item embeddings.
    \item Through theoretical analysis, we reveal that the generalization problem of item embedding learning in the multi-task learning framework can be effectively addressed using sharpness-aware minimization.
    \item Based on this insight, we propose a novel framework, \method{}, which incorporates sharpness-aware minimization into both local training and global aggregation to stabilize the training process and improve the generalization of embeddings.
    \item We conduct extensive experiments on four real-world datasets and validate the effectiveness of \method{}.
    Our method, \method{}, consistently outperforms all baselines across diverse scenarios.
    Its advantage grows as the user-item ratio increases.
\end{itemize}

%% file: body/02_related_works.tex
\section{Related Work}


\subsection{Federated Recommendation}
Recommendation systems~\cite{isinkaye2015recommendation, zhang2019deep, ko2022survey, feng2025plug, feng2025raid} are designed to leverage user-item interaction information to provide personalized content to users~\cite{he2017neural}.
In federated recommendation, each user is treated as a client, holding their own interaction data locally~\cite{jiang2025tutorial, chen2024confusion, ke2025early}.
Existing research on federated recommender systems generally falls into two categories.
First, a number of studies focus on clustering similar users based on user embeddings or item embeddings updated by clients, and then conducting collaborative training within each cluster~\cite{he2024co, luo2022personalized, mao2024cluster, muhammad2020fedfast, luo2024perfedrec++}.
These approaches aim to stabilize the training process by grouping clients with relatively small data distributional differences, thereby alleviating the effects of data heterogeneity and sparsity within clusters.
However, obtaining reliable clustering results is challenging due to the limited and sparse representations of individual client data.
Second, other studies aim to train personalized models for each client by adapting the global model through local fine-tuning~\cite{zhang2023dual, zhang2024gpfedrec, xiao2025points, zhang2024cf, feng2024controllable} or by fusing the global and local models~\cite{li2023federated, li2025personalized, lu2025dammfnd, wang2024computing, feng2025survey}.
These methods allocate more model parameters to capture client-specific data distribution, thereby mitigating the negative impact of heterogeneity and sparsity across clients.

Both categories of methods overlook a \textbf{critical challenge}, i.e., the stable learning of a generalized item embedding throughout the federated recommender system training process.
Unlike existing approaches, our proposed \method{} directly learn generalized embedding, thereby enhancing recommendation performance.

\subsection{Sharpness Aware Minimization}
The training loss landscapes of today's deep learning models are commonly complex and non-convex, containing a multitude of local and global minima~\cite{wu2017towards, ouyang2024learn}.
Motivated by studies on the correlation between the geometry of the loss landscape and model generalization, ~\citet{foret2021sharpness} introduced sharpness-aware minimization (SAM), an efficient training scheme~\cite{rostand2024comprehensive, pillarhist}.
SAM aims to identify parameters that lie within neighborhoods of uniformly low loss value by minimizing the worst-case loss in a neighborhood of the current model parameter $\boldsymbol{\theta}$, given by:
\begin{equation}
    \min_{\boldsymbol{\theta}} \max_{\|\boldsymbol{\epsilon}\|_2 \leq \rho} \mathcal{L}(\boldsymbol{\theta} + \boldsymbol{\epsilon}),
    \label{eq:sam}
\end{equation}
where $\|\cdot\|_2$ denotes the $l_2$ norm and $\rho$ represents the radius of the neighborhood.

To solve problem~\eqref{eq:sam}, researchers first address the inner maximization, yielding:
\begin{equation}
\boldsymbol{\epsilon}^* = \rho \frac{\nabla_{\boldsymbol{\theta}} \mathcal{L}(\boldsymbol{\theta})}{\|\nabla_{\boldsymbol{\theta}} \mathcal{L}(\boldsymbol{\theta})\|_2}.
\end{equation}
Subsequently, the gradient with respect to this perturbed model is computed to update $\boldsymbol{\theta}$:
\begin{equation}
\mathbf{g}^{\mathrm{SAM}} = \nabla_{\boldsymbol{\theta}} \left[ \max_{\|\boldsymbol{\epsilon}\| \leq \rho} \mathcal{L}(\boldsymbol{\theta} + \boldsymbol{\epsilon}) \right] \approx \nabla_{\boldsymbol{\theta}} \mathcal{L}(\boldsymbol{\theta} + \boldsymbol{\epsilon}^*).
\label{eq:sam_gradient}
\end{equation}
This ensures the model converges to flatter minima, which improves generalization.

The detailed derivation is provided in Appendix~\ref{sec:sam_solution}.

%% file: body/03_preliminary.tex
\section{Preliminaries}

\subsection{Problem Statement}
Let $\mathcal{U} = \left\{u_1, u_2, \dots, u_m\right\}$ denote the user set with $m$ users and $\mathcal{I} = \left\{i_1, i_2, \dots, i_n\right\}$ denote the item set with $n$ items.
In federated recommender systems, each user is treated as an individual client, which locally stores their interaction records $\tilde{\mathcal{R}}_{k} = \left[\tilde{r}_{k1}, \tilde{r}_{k2}, \dots, \tilde{r}_{km}\right]$, where $\tilde{r}_{kl} = 1$ if user $u_k$ has interacted with item $i_l$, and $\tilde{r}_{kl} = 0$ otherwise.
These observed interaction records are sampled from the underlying set $\mathcal{R}_k = \left[r_{k1}, r_{k2}, \dots, r_{km}\right]$, where some user-item interactions may be unobserved in $\tilde{\mathcal{R}}_k$ because user has not encountered the corresponding items.

The recommendation model deployed on client $k$ consists of a user embedding $\boldsymbol{u}_{k} \in \mathbb{R}^{d}$, a score function parameterized by $\boldsymbol{\theta}^{\mathrm{score}}_k$, and item embeddings $\boldsymbol{\theta}^{\mathrm{item}}_{k} \in \mathbb{R}^{n \times d}$.
During communication round $t$, only the score function and item embeddings, denoted as $\boldsymbol{\theta}^{\mathrm{share}}_{k} = \{\boldsymbol{\theta}^{\mathrm{score}}_{k}, \boldsymbol{\theta}^{\mathrm{item}}_{k}\}$, are uploaded to the central server.
The server then performs aggregation to obtain the global parameters $\boldsymbol{\theta}^{\mathrm{global}}_{t} = \{\boldsymbol{\theta}^{\mathrm{score}}_{t}, \boldsymbol{\theta}^{\mathrm{item}}_{t}\}$, which are distributed back to the clients for the next communication round.

The objective is to collaboratively train recommendation models to predict user-item ratings across decentralized clients, while preserving privacy by avoiding the exchange of raw interaction records or user embeddings. 
Formally, the optimization problem is defined as:
\begin{equation}
    \min \sum_{k=1}^{m} \mathbb{E}_{i \sim \mathcal{I}} \left[\mathcal{L}_{\tilde{\mathcal{R}}_k} \left(\boldsymbol{u}_{k}, \boldsymbol{\theta}^{\mathrm{score}}_k, \boldsymbol{\theta}^{\mathrm{item}}_k\right)\right]
\end{equation}
where $\mathcal{L}_{\tilde{\mathcal{R}}_k}$ denotes the loss function on $\tilde{\mathcal{R}}_k$.

\subsection{Problem Reformulating in Item-Centered Perspective}
Existing work has formalized the federated recommendation problem from a user-centered perspective, as discussed above.
In this section, we reformulate this problem in an item-centered perspective and cast it as a multi-task learning problem.

For item $i_l$, the item-user interaction record is denoted as $\mathcal{S}_{l} = \left[\tilde{r}_{1l}, \tilde{r}_{2l}, \dots, \tilde{r}_{ml}\right]$, which is sampled from the underlying set $\mathcal{D}_{l} = \left[r_{1l}, r_{2l}, \dots, r_{ml}\right]$.
A prediction task $k$ associate with item $i_l$ corresponds to determining whether item $i_l$ is interacted with by user $u_k$.
We denote the empirical loss for task $k$ associate with item $i_l$ as $\mathcal{L}_{\tilde{r}_{kl}}(\boldsymbol{u}_k, \boldsymbol{\theta}^{\mathrm{score}}_k, \boldsymbol{\theta}^{\mathrm{item}}_k(l))$, where $\boldsymbol{\theta}^{\mathrm{item}}_k(l)$ represents the embedding of item $i_l$.
For simplicity, we denote $\boldsymbol{\theta}^{kl} = \{\boldsymbol{u}_k, \boldsymbol{\theta}^{\mathrm{score}}_k, \boldsymbol{\theta}^{\mathrm{item}}_k(l)\}$, and abbreviate the empirical loss above as $\mathcal{L}_{\mathcal{S}_l}(\boldsymbol{\theta}^{kl})$.
Similarly, the general loss for task $k$ associate with item $i_l$ is denoted as $\mathcal{L}_{\mathcal{D}_l}(\boldsymbol{\theta}^{kl})$.
In the following discussion, we focus a specific item $i_l$, and omit the index $l$ whenever it does not cause ambiguity.

With these notations, the objective of the federated recommendation problem for a specific item $i_l$ is to simultaneously minimize the general losses for all tasks:
\begin{equation}
    \min_{\boldsymbol{\theta}^{kl}, k \in [m]} \left[
        \mathcal{L}_{\mathcal{S}_l}\left(\boldsymbol{\theta}^{1l}\right), \mathcal{L}_{\mathcal{S}_l}\left(\boldsymbol{\theta}^{2l}\right), \dots, \mathcal{L}_{\mathcal{S}_l}\left(\boldsymbol{\theta}^{ml}\right)
    \right],
\end{equation}
By computing the gradient $\boldsymbol{g}^{k}$ for the $k$-th task ($k \in [m]$), the model parameters are updated using a unified gradient $\boldsymbol{g} = \mathrm{Agg}([\boldsymbol{g}^k]_{k=1}^{m})$, where $\mathrm{Agg(\cdot)}$ denotes a generic aggregation operator that combines task-specific gradients, as commonly employed in gradient-based multi-task learning.

Accordingly, the overall objective of the federated recommendation problem is:
\begin{equation}
    \min_{\left\{\boldsymbol{\theta}^{kl}, k \in [m]\right\}} \sum_{l=1}^{n}
    \left[
        \mathcal{L}_{\mathcal{S}_l}\left(\boldsymbol{\theta}^{1l}\right), \mathcal{L}_{\mathcal{S}_l}\left(\boldsymbol{\theta}^{2l}\right), \dots, \mathcal{L}_{\mathcal{S}_l}\left(\boldsymbol{\theta}^{ml}\right)
    \right].
\end{equation}

%% file: body/04_method.tex
\section{Methodology}
Existing methods primarily focus on minimizing empirical losses and do not explicitly consider the general loss, which is crucial for improving the generalization ability of the model.
Since we only have access to the observed training set, it is challenging to directly enhance the generalization of the learned embeddings.

Through the theoretical analysis in section~\ref{sec:theo_analy}, we establish a connection between improving generalization and minimizing empirical loss via SAM.
First, we leverage Gaussian-perturbed PAC-Bounds~\cite{mcallester1999pac, alquier2016properties, lu2024generic} to relate the loss on the true distribution to the loss on the empirical distribution.
Next, we reformulate the minimization of the expected empirical loss as an upper bound in the form of SAM.
Furthermore, to incorporate multi-task learning~\cite{zhang2018overview, zhang2021survey}, we introduce a hierarchical multi-head model architecture, consisting of a shared set of parameters and a group of task-specific private parameters.

Building on this theoretical analysis, we propose \method{}, which integrates SAM into both local training and global aggregation.
In the implementation, the shared parameters correspond to those uploaded to the server and aggregated across clients, while the private parameters correspond to the user-specific embeddings that remain local.
The computational implementation and training algorithm are presented in Section~\ref{sec:comp_imple}.

\subsection{Theoretical Analysis}\label{sec:theo_analy}
In the following discussion, we focus on a specific item $i_l$ and omit the index $l$ in the formulas for simplicity.

\subsubsection{Relating the General Loss to the Empirical Loss}
The objective of improving the generalization of the item embeddings is to minimize the loss with respect to the true distribution $D$, which is typically unknown.
In practice, however, we only have access to the empirical distribution $\mathcal{S}$.
A feasible approach to bridge the gap between the loss on the true distribution and the empirical loss is to leverage Gaussian-perturbed PAC-Bounds.

\begin{lemma}[Multi Gaussian-Perturbed PAC Bound]
    With the assumption that adding a Gaussian perturbation will raise the test error: $\mathcal{L}_{\mathcal{D}}(\boldsymbol{\theta}^k) \le \mathbb{E}_{\boldsymbol{\varepsilon}\sim\mathcal{N}(0,\sigma^2 \mathbb{I})}\bigl[\mathcal{L}_{\mathcal{D}}(\boldsymbol{\theta}^k+\boldsymbol{\varepsilon})\bigr]$.
    Let $T_k$ be the number of parameters in $\boldsymbol{\theta}^k$ and $N$ be the cardinality of $\mathcal{S}$. 
    Then, with probability  $1-\gamma$ (over the choice of training set $\mathcal{S} \sim \mathcal{D}$), the following holds:
    \begin{equation}
        \bigl[\mathcal{L}_{\mathcal{D}} (\boldsymbol{\theta}^k) \bigr]_{k=1}^m \le \left[\mathbb{E}_{\boldsymbol{\varepsilon} \sim \mathcal{N}(0,\sigma^2 \mathbb{I})} \mathcal{L}_{\mathcal{S}}(\boldsymbol{\theta}^k + \boldsymbol{\varepsilon}) + f^k\left(\|\boldsymbol{\theta}^k\|_2^2\right)\right]_{k=1}^m,
        \label{eq:multi-gau-pac-bound}
    \end{equation}
    where $f^k\left(\|\boldsymbol{\theta}^k\|_2^2\right)$ is a regularization term, equals to
    \begin{equation}
        \frac{1}{\sqrt{N}} \Biggl[ \dfrac{1}{2} + \dfrac{T_k}{2}\log \left(1+\dfrac{\|\boldsymbol{\theta}^k\|_2^2}{T_k\sigma^2}\right) + \log\dfrac{1}{\gamma} + 6\log(N+T_k) + \dfrac{L^2}{8} \Biggr],
    \end{equation}
    and $L$ is an upper bound on the loss function.
    
    \label{lem:multi-gau-pac-bound}
    
    \begin{proof}
        %
        The detailed proof is given in Appendix~\ref{sec:proof_lem_1}.
    \end{proof}
\end{lemma}

Through Lemma~\ref{lem:multi-gau-pac-bound}, we establish a connection between generalization improvement and empirical risk minimization.
Specifically, minimizing the loss on the true distribution can be reformulated as minimizing the empirical loss augmented with a regularization term.
This result directly relates the loss on the true distribution to that on the empirical distribution.

\subsubsection{From Empirical Loss Minimization to SAM}
The empirical loss in Lemma~\ref{lem:multi-gau-pac-bound} corresponds to the expected loss under Gaussian perturbations.
In the following, we explain how to transition from the PAC-Bayes bound with Gaussian perturbations to a worst-case perturbation bound, i.e., SAM.
In doing so, we transform the generalization problem into a SAM problem, and subsequently employ SAM within our proposed \method{} framework to address generalization.
To accommodate multi-task learning scenarios, we adopt a hierarchically multi-head network architecture consisting of a shared set of parameters and a set of task-specific parameters.
This design is effective for handling multiple tasks and naturally aligns with the cross-device setting in federated recommendation: the shared set of parameters corresponds to those uploaded to and aggregated by the server at each round, while the task-specific parameters correspond to user embeddings that remain local on the clients.

Specifically, we consider $m$ tasks (users).
At a communication round $t$, for task $k$, we split the parameters into a shared part $\left\{\boldsymbol{\theta}^{\mathrm{score}}_t, \boldsymbol{\theta}^{\mathrm{item}}_t\right\} = \boldsymbol{\theta}^{\mathrm{global}}_t = \boldsymbol{\theta}_{\mathrm{co}} \in \mathbb{R}^{T_{\mathrm{co}}}$ and a task-specific part $\boldsymbol{u}_k = \boldsymbol{\theta}^{k}_{\mathrm{ur}} \in \mathbb{R}^{T_{\mathrm{ur}}}$, so that the overall parameter vector is given by $\boldsymbol{\theta}^k = (\boldsymbol{\theta}_{\mathrm{co}}, \boldsymbol{\theta}_{\mathrm{ur}}^k) \in \mathbb{R}^{T}$, $T = T_{\mathrm{co}} + T_{\mathrm{ur}}$.
With this formulation, the empirical loss minimization can be reformulated in the SAM framework.

\begin{lemma}[Hierarchical SAM]
    For any perturbation radii $\rho_{\mathrm{co}},\rho_{\mathrm{ur}}>0$, with probability $1-\gamma$ (over the choice of training set $\mathcal{S} \sim \mathcal{D}$) we obtain:
    \begin{equation}
        \begin{aligned}
            \bigl[\mathcal{L}_{\mathcal{D}}(\boldsymbol{\theta}^k)\bigr]_{k=1}^m
            &\le \max_{\|\boldsymbol{\varepsilon}_{\mathrm{co}}\|_2 \le \rho_{\mathrm{co}}} 
            \max_{\|\boldsymbol{\varepsilon}_{\mathrm{ur}}^k\|_2 \le \rho_{\mathrm{ur}}} \\
            &\quad \left[\mathcal{L}_{\mathcal{S}}(\boldsymbol{\theta}_{\mathrm{co}} + \boldsymbol{\varepsilon}_{\mathrm{co}},
            \boldsymbol{\theta}_{\mathrm{ur}}^k + \boldsymbol{\varepsilon}_{\mathrm{ur}}^k)
            + f^k\left(\|\boldsymbol{\theta}^k\|_2^2\right)\right]_{k=1}^m,
        \end{aligned}
        \label{eq:hie-sam}
    \end{equation}
    where $f^k(\|\boldsymbol{\theta}^k\|_2^2)$ is defined the same as in Lemma~\ref{lem:multi-gau-pac-bound}.
    
    \label{lem:hie-sam}
    
    \begin{proof}
        %
        The detailed proof is given in Appendix~\ref{sec:proof_lem_2}.
    \end{proof}
\end{lemma}

According to Lemma~\ref{lem:hie-sam}, the generalization objective in multi-task learning can be reformulated as follows:
\begin{equation}
    \begin{aligned}
        & \min_{\boldsymbol{\theta}_\mathrm{co},\boldsymbol{\theta}_\mathrm{ur}^{1:m}} 
          \max_{\|\boldsymbol{\varepsilon}_\mathrm{co}\|_2\le\rho_\mathrm{co}} 
          \max_{\|\boldsymbol{\varepsilon}_\mathrm{ur}^k\|_2\le\rho_\mathrm{ur}}
          \bigl[\mathcal{L}_{\mathcal{S}}(\boldsymbol{\theta}_\mathrm{co} +\boldsymbol{\varepsilon}_\mathrm{co},
          \boldsymbol{\theta}_\mathrm{ur}^k+\boldsymbol{\varepsilon}_\mathrm{ur}^k)
          + f^k(\|\boldsymbol{\theta}^k\|_2^2)\bigr]_{k=1}^m \\
        =\, & \underbrace{\min_{\boldsymbol{\theta}_\mathrm{co},\boldsymbol{\theta}_\mathrm{ur}^{1:m}} 
          \left[ \max_{\|\boldsymbol{\varepsilon}_\mathrm{co}\|_2\le\rho_\mathrm{co}} 
          \max_{\|\boldsymbol{\varepsilon}_\mathrm{ur}^k\|_2\le\rho_\mathrm{ur}} 
          \mathcal{L}_{\mathcal{S}}(\boldsymbol{\theta}_\mathrm{co} +\boldsymbol{\varepsilon}_\mathrm{co},
          \boldsymbol{\theta}_\mathrm{ur}^k+\boldsymbol{\varepsilon}_\mathrm{ur}^k)\right]_{k=1}^m}_{\displaystyle\mathrm{SAM}} \\
        &\quad + \left[f^k\left(\|\boldsymbol{\theta}^k\|_2^2\right)\right]_{k=1}^m \\
        =\, & \underbrace{\min_{\boldsymbol{\theta}_\mathrm{co},\boldsymbol{\theta}_\mathrm{ur}^{1:m}}  
          \max_{\|\boldsymbol{\varepsilon}_\mathrm{co}\|_2\le\rho_\mathrm{co}} 
          \left[ \max_{\|\boldsymbol{\varepsilon}_\mathrm{ur}^k\|_2\le\rho_\mathrm{ur}} 
          \mathcal{L}_{\mathcal{S}}^k(\boldsymbol{\theta}_\mathrm{co} + \boldsymbol{\varepsilon}_\mathrm{co}, 
          \boldsymbol{\theta}_\mathrm{ur}^k+\boldsymbol{\varepsilon}_\mathrm{ur}^k) \right]_{k=1}^m}_{\displaystyle\mathrm{SAM}} \\
        &\quad + \left[f^k\left(\|\boldsymbol{\theta}^k\|_2^2\right)\right]_{k=1}^m.
    \end{aligned}
    \label{eq:mtl-sam}
\end{equation}
Since the parameter regularization term $f^k(\|\boldsymbol{\theta}^k\|_2^2)$ is independent of the perturbation $\boldsymbol{\varepsilon}_\mathrm{co}^k$ and $\boldsymbol{\varepsilon}_\mathrm{ur}^k$, it can be moved outside the inner maximization.
At this point, the outer objective in Eq~\eqref{eq:mtl-sam} consists of two components: the first is a standard SAM problem, and the second is the regularization term.

\begin{algorithm}[t]
    \caption{Training Procedure of \method{}}
    \label{alg:method-algorithm}
    \begin{algorithmic}[1]
        \Require $T, n_{\mathrm{sample}}, \rho_{\mathrm{co}}, \rho_{\mathrm{ur}}, \eta$
        \Statex \hspace*{-\algorithmicindent} \textbf{Initialize:} $\boldsymbol{\theta}^{\mathrm{global}}_0$
        
        \Statex \hspace*{-\algorithmicindent} \textsc{Global Procedure}:
        \setcounter{ALG@line}{0}
        \ForAll{client index $k = 1, 2, \ldots, m$ \textbf{in parallel}}
            \State initialize $\boldsymbol{u}_k, \boldsymbol{\theta}^{\mathrm{share}}_k$
        \EndFor
        \For{$t = 1$ to $T$}
            \State $\textrm{client list} \gets$ randomly select $n_{\mathrm{sample}}$ clients from $m$ clients
            \ForAll{$k \in \textrm{client list}$ \textbf{in parallel}}
                \State $\boldsymbol{g}^{k, \mathrm{SAM}}_{\mathrm{co}} \gets \Call{ClientUpdate}{k, \boldsymbol{\theta}^{\mathrm{global}}_{t-1}}$
            \EndFor
            \State $\boldsymbol{g} \gets \frac{1}{n_{\mathrm{sample}}}\sum_{k=1}^{n_{\mathrm{sample}}} \boldsymbol{g}^{k, \mathrm{SAM}}_{\mathrm{co}}$ \Comment{Server aggregation}
            \State $\boldsymbol{\theta}^{\mathrm{global}}_t \gets \boldsymbol{\theta}^{\mathrm{global}}_{t-1} - \eta \boldsymbol{g}$
        \EndFor
        \State \Return $\boldsymbol{u}_{1:m}, \boldsymbol{\theta}^{\mathrm{global}}_{T}$
        
        \Statex
        \Statex \hspace*{-\algorithmicindent} \textsc{ClientUpdate}$(k, \boldsymbol{\theta}^{\mathrm{global}})$:
        \setcounter{ALG@line}{0}
        \State sample negative items at a 1:4 ratio with positive items
        \State compute $\boldsymbol{g}_{\mathrm{norm}}$ using Eq~\eqref{eq:mtl-sam}
        \State compute $\boldsymbol{\varepsilon}^{k,*}_{\mathrm{ur}}$ and $\boldsymbol{g}^{k, \mathrm{SAM}}_{\mathrm{ur}}$ using Eq~\eqref{eq:pert-non-share}~\eqref{eq:grad-non-share}
        \State update the $\boldsymbol{u}_k$ with $\boldsymbol{g}^{k, \mathrm{SAM}}_{\mathrm{ur}} + \boldsymbol{g}_{\mathrm{norm}}$ using Eq~\eqref{eq:update-non-shared}
        \State compute $\boldsymbol{\varepsilon}^{k,*}_{\mathrm{co}}$ and $\boldsymbol{g}^{k,\mathrm{SAM}}_{\mathrm{co}}$ using Eq~\eqref{eq:pert-share}~\eqref{eq:grad-share}
        \State \Return $\boldsymbol{g}^{k,\mathrm{SAM}}_{\mathrm{co}} + \boldsymbol{g}_{\mathrm{norm}}$
    \end{algorithmic}
\end{algorithm}

\subsection{Computational Implementation}\label{sec:comp_imple}
As two terms in Eq~\eqref{eq:mtl-sam} are additive and the gradient of the regularization term with respect to the parameters $\boldsymbol{\theta}$ can be computed directly, we focus primarily on optimizing the SAM component.
Due to the hierarchical structure of the model, the parameter gradients are also hierarchical.
Therefore, the updates of the shared parameters $\boldsymbol{\theta}_{\mathrm{co}}$ and the private parameters $\boldsymbol{\theta}_{\mathrm{ur}}$ can be decoupled.
Based on Eq~\eqref{eq:sam_gradient}, the update rule can be expressed as follows.

For the optimization of the SAM component, we can adopt an initial and straightforward update approach, which can be divided into two specific steps.

\paragraph{Update the non-shared parts}
Since the non-shared perturbations $\boldsymbol{\varepsilon}^{k}_{\mathrm{ur}}, k \in [m]$ are independent to each task, for task $k$, we update its non-shared part $\boldsymbol{\theta}_{\mathrm{ur}}^k$:
\begin{align}
    & \boldsymbol{\varepsilon}^{k,*}_{\mathrm{ur}} = \rho_{\mathrm{ur}} \frac{\nabla_{\boldsymbol{\theta}_{\mathrm{ur}}^k} \mathcal{L}_\mathcal{S}(\boldsymbol{\theta}_{\mathrm{co}}, \boldsymbol{\theta}_{\mathrm{ur}}^k)}{\left\|\nabla_{\boldsymbol{\theta}_{\mathrm{ur}}^k} \mathcal{L}_\mathcal{S}(\boldsymbol{\theta}_{\mathrm{co}}, \boldsymbol{\theta}_{\mathrm{ur}}^k) \right\|_2} \label{eq:pert-non-share} \\
    \quad & \boldsymbol{g}^{k, \mathrm{SAM}}_{\mathrm{ur}} = \nabla_{\boldsymbol{\theta}_{\mathrm{ur}}^k} \mathcal{L}_\mathcal{S}(\boldsymbol{\theta}_{\mathrm{co}}, \boldsymbol{\theta}_{\mathrm{ur}}^k + \boldsymbol{\varepsilon}^{k}_{\mathrm{ur}}) \label{eq:grad-non-share} \\
    \quad & \boldsymbol{\theta}_{\mathrm{ur}}^k = \boldsymbol{\theta}_{\mathrm{ur}}^k - \eta \boldsymbol{g}^{k, \mathrm{SAM}}_{\mathrm{ur}},
    \label{eq:update-non-shared}
\end{align}
where $\boldsymbol{\varepsilon}^{k,*}_{\mathrm{ur}}$ denotes the worst-case perturbation on the $k$-th private task parameter, and $\eta > 0$ represents the learning rate.
    
\paragraph{Update the shared part}
Updating the shared part $\boldsymbol{\theta}_{\mathrm{co}}$ is more challenging because its worst-cased perturbation $\boldsymbol{\varepsilon}_{\mathrm{co}}$ is shared among the tasks. 
To derive how to update $\boldsymbol{\theta}_{\mathrm{co}}$ with respect to all tasks, we first discuss the case when we update this with respect to task $k$ without caring about other tasks. 
Specifically, this task's SAM shared gradient is computed as:
\begin{align}
    & \boldsymbol{\varepsilon}^{k,*}_{\mathrm{co}} = \rho_{\mathrm{co}} \frac{\nabla_{\boldsymbol{\theta}_{\mathrm{co}}} \mathcal{L}_\mathcal{S}(\boldsymbol{\theta}_{\mathrm{co}}, \boldsymbol{\theta}_{\mathrm{ur}}^k)}{\left\| \nabla_{\boldsymbol{\theta}_{\mathrm{co}}} \mathcal{L}_\mathcal{S}(\boldsymbol{\theta}_{\mathrm{co}}, \boldsymbol{\theta}_{\mathrm{ur}}^k) \right\|_2} \label{eq:pert-share} \\
    & \boldsymbol{g}^{k,\mathrm{SAM}}_{\mathrm{co}} = \nabla_{\boldsymbol{\theta}_{\mathrm{co}}} \mathcal{L}_\mathcal{S}(\boldsymbol{\theta}_{\mathrm{co}} + \boldsymbol{\varepsilon}^{k,*}_{\mathrm{co}}, \boldsymbol{\theta}_{\mathrm{ur}}^k), \label{eq:grad-share}
\end{align}
then we have a straightforward updating strategy:
\begin{equation*}
    \boldsymbol{g}^{\mathrm{SAM}}_{\mathrm{co}} = \mathrm{Agg}(\boldsymbol{g}^{1,\mathrm{SAM}}_{\mathrm{co}}, \ldots, \boldsymbol{g}^{m,\mathrm{SAM}}_{\mathrm{co}})
    \quad \Rightarrow{} \quad
    \boldsymbol{\theta}_{\mathrm{co}} = \boldsymbol{\theta}_{\mathrm{co}} - \eta \boldsymbol{g}^{\mathrm{SAM}}_{\mathrm{co}}.
\end{equation*}
Here, $\textrm{Agg}(\cdot)$ denotes the gradient aggregation operator.
For computational efficiency and practical applicability, we adopt the widely used FedAvg-style aggregation~\cite{mcmahan2017communication, xu2024fakeshield, lu2022understanding}, i.e., the global gradient is obtained as the weighted average of client gradients.

\paragraph{Implementation}
Putting everything together, we address a federated learning setting that involves multiple multi-task learning problems.
For computational efficiency, we do not compute the worst-case perturbation and SAM gradient for each item individually.
Instead, following the standard training paradigm in federated learning, we sample a subset of clients and perform training on them.
The computed worst-case perturbation can then be regarded as the worst-case perturbation across the multi-task learning setting, and the resulting SAM gradient corresponds to the aggregated gradient over these tasks.
This strategy strikes a balance between computational tractability and generalization performance.

The detailed steps of our proposed method are summarized in Algorithm~\ref{alg:method-algorithm}.

%% file: body/05_experiment.tex
\section{Experiments}
To comprehensively evaluate our proposed method, we conduct experiments on four real-world datasets.
Specifically, we aim to answer the following Research Questions (RQs):
\begin{itemize}[leftmargin=*]\setlength{\itemsep}{-\itemsep}
    \item \textbf{RQ1}: How does our method perform on recommendation tasks compared with strong baselines?
    \item \textbf{RQ2}: How do the principal hyperparameters, $\rho_{\mathrm{ur}}$ and $\rho_{\mathrm{co}}$, affect \method{}, thereby influencing recommendation performance?
    \item \textbf{RQ3}: What are the respective contributions of SAM training on the non-shared parts and on the shared parts?
    \item \textbf{RQ4}: We visualize the loss landscape to determine whether \method{} converges to a flatter, more generalizable solution.
\end{itemize}

\subsection{Experimental Settings}
\paragraph{Datasets.}
\begin{table}[h]
\centering
\caption{Dataset statistics.}
\label{tab:dataset}
\resizebox{\linewidth}{!}{
    \begin{tabular}{ccccc}
        \toprule
        Dataset & \makecell[c]{\# Users \\ (\# Clients)} & \# Items & \# Interactions & Sparsity \\ \midrule
        FilmTrust & 1,227 & 2,059 & 34,889 & 98.62\% \\
        Lastfm-2K & 1,600 & 12,454 & 185,650 & 99.07\% \\
        Amazon-Video & 8,072 & 11,830 & 63,836 & 99.93\% \\
        QB-article & 24,516 & 7,355 & 348,736 & 99.81\% \\
        \bottomrule
    \end{tabular}
}
\end{table}
We conduct experiments on four publicly available real-world recommendation datasets, adapting them to federated recommendation scenarios.
In the cross-device setting, each user acts as a client, maintaining their interaction records locally.
Specifically, FilmTrust~\cite{guo2016novel} is a movie recommendation dataset.
Lastfm-2K~\cite{cantador2011second} is a music dataset, where each user retains a list of listened artists along with their listening frequency.
Amazon-Video~\cite{ni2019justifying} consists of product reviews and metadata collected from the Amazon website.
QB-article~\cite{yuan2022tenrec} records user click behaviors on articles.
The overall dataset statistics are shown in Table~\ref{tab:dataset}.

\paragraph{Data Preprocessing}
For all datasets, we remove the users with less than 5 interactions.
Following~\cite{zhang2023dual}, we randomly sample $N=4$ negative instances for each positive sample during training.
We employ the leave-one-out strategy.
The most recent interaction item for each user (sorted by interaction timestamp) is retained for testing.
For each user, we randomly select 99 unobserved items and perform a ranking evaluation among 100 items, including the test item.

\paragraph{Baselines}
To assess \method{} and demonstrate its superiority, we choose the following algorithms as baselines:
\begin{itemize}[leftmargin=*]\setlength{\itemsep}{-\itemsep}
    \item \textbf{FedNCF}~\cite{perifanis2022federated}: This is the federated adaptation of neural collaborative filtering~\cite{he2017neural}. It updates user embedding locally on each client and synchronizes the item embedding globally on the server.
    \item \textbf{FedMF}~\cite{chai2020secure}: It applies matrix factorization in the federated learning environment to prevent information leakage by encrypting gradients of both user and item embeddings.
    \item \textbf{PerFedRec}~\cite{luo2022personalized}: It integrates a federated GNN for joint representation learning, user clustering, and model adaptation.
    \item \textbf{PFedRec}~\cite{zhang2023dual}: It incorporates the user embeddings into the score function and performs local adaptation to generate personalized item representations for each user.
    \item \textbf{FedRAP}~\cite{li2023federated}: It learns a global view of items via federated learning and a personalized view locally on each user, and enforces the two views to be complementary.
    \item \textbf{CoFedRec}~\cite{he2024co}: This approach employs a co-clustering strategy, first grouping items into clusters and then clustering users based on their preference for specific item categories.
    \item \textbf{GPFedRec}~\cite{zhang2024gpfedrec}: It constructs a user relationship graph using the received item embeddings and learns user-specific item embeddings through graph-guided aggregation.
\end{itemize}

\paragraph{Evaluation Metrics}
Model performance is evaluated using Tok-K evaluation metrics~\cite{he2015trirank}, including Hit Ratio (HR) and Normalized Discounted Cumulative Gain (NDCG).
HR@K measures whether the test item is in the top-K list. NDCG@K is a position-aware ranking metric that gives higher scores to hits that occur at higher ranks.

\begin{table*}[h]
\centering
\caption[Recommendation performance]{Results of recommendation performance. Best results are typeset in bold with a colored background, e.g., \protect\colorbox[HTML]{95D4EE}{\textbf{89.38$\pm$0.21}}. Runner-up results use a different background color, e.g., \protect\colorbox[HTML]{C8EBF6}{66.67$\pm$1.09}. We run all models 5 times and report the average results and standard deviation. Results are expressed as percentages (\%).}

\label{tab:main_results}
\begin{tabular}{ccccccccc}
    \toprule
    Dataset & \multicolumn{4}{c}{FilmTrust} & \multicolumn{4}{c}{Lastfm-2K} \\ \cmidrule(lr){2-5} \cmidrule(lr){6-9}
    Methods & HR@5 & NDCG@5 & HR@10 & NDCG@10 & HR@5 & NDCG@5 & HR@10 & NDCG@10 \\ \midrule
    FedNCF & 66.67$\pm$0.30 & 53.09$\pm$0.36 & 69.49$\pm$0.17 & 54.20$\pm$0.39 & 52.08$\pm$0.74 & 38.98$\pm$0.28 & 63.85$\pm$0.49 & 42.82$\pm$0.26 \\
    FedMF & 65.23$\pm$0.30 & 47.83$\pm$0.35 & 69.38$\pm$0.45 & 49.21$\pm$0.40 & 73.10$\pm$0.28 & 65.02$\pm$0.16 & 78.63$\pm$0.18 & 66.71$\pm$0.23 \\
    PerFedRec & \cellcolor[HTML]{C8EBF6} 81.77$\pm$1.64 & 66.35$\pm$1.67 & 90.52$\pm$0.91 & 76.95$\pm$1.71 & 45.94$\pm$1.07 & 37.83$\pm$2.13 & 63.96$\pm$3.70 & 44.24$\pm$3.79 \\
    PFedRec & 66.40$\pm$0.33 & 54.30$\pm$0.39 & 78.76$\pm$0.14 & 64.83$\pm$0.36 & \cellcolor[HTML]{95D4EE} \textbf{76.40$\pm$0.45} & \cellcolor[HTML]{C8EBF6}  70.59$\pm$0.21 & \cellcolor[HTML]{95D4EE} \textbf{81.35$\pm$0.62} & \cellcolor[HTML]{C8EBF6} 72.34$\pm$0.12 \\
    FedRAP & 87.83$\pm$0.77 & \cellcolor[HTML]{C8EBF6} 73.80$\pm$1.73 & \cellcolor[HTML]{C8EBF6} 91.74$\pm$0.10 & \cellcolor[HTML]{C8EBF6} 77.67$\pm$1.29 & 68.33$\pm$0.91 & 64.55$\pm$0.69 & 75.71$\pm$0.31 & 71.53$\pm$0.24 \\
    CoFedRec & 68.87$\pm$0.35 & 55.58$\pm$0.53 & 81.26$\pm$0.24 & 67.29$\pm$0.09 & 73.85$\pm$0.94 & 68.68$\pm$0.55 & 77.23$\pm$1.23 & 68.96$\pm$0.97 \\
    GPFedRec & 66.56$\pm$0.57 & 53.32$\pm$0.79 & 78.57$\pm$0.13 & 64.07$\pm$0.21 & 70.92$\pm$1.58 & 61.41$\pm$2.57 & 80.35$\pm$0.40 & 70.70$\pm$1.01 \\
    \method{} & \cellcolor[HTML]{95D4EE} \textbf{89.38$\pm$0.21} & \cellcolor[HTML]{95D4EE} \textbf{81.33$\pm$0.32} & \cellcolor[HTML]{95D4EE} \textbf{91.66$\pm$0.10} & \cellcolor[HTML]{95D4EE} \textbf{82.01$\pm$0.23} & \cellcolor[HTML]{C8EBF6} 73.92$\pm$0.22 & \cellcolor[HTML]{95D4EE} \textbf{70.79$\pm$0.16} & \cellcolor[HTML]{C8EBF6} 80.69$\pm$0.25 & \cellcolor[HTML]{95D4EE} \textbf{73.56$\pm$0.14} \\
    \bottomrule
    Dataset & \multicolumn{4}{c}{Amazon-Video} & \multicolumn{4}{c}{QB-article} \\ \cmidrule(lr){2-5} \cmidrule(lr){6-9}
    Methods & HR@5 & NDCG@5 & HR@10 & NDCG@10 & HR@5 & NDCG@5 & HR@10 & NDCG@10 \\ \midrule
    FedNCF & \cellcolor[HTML]{C8EBF6} 48.79$\pm$1.09 & 35.04$\pm$1.38 & \cellcolor[HTML]{C8EBF6} 60.88$\pm$0.34 & 38.86$\pm$1.30 & 32.58$\pm$0.04 & 19.79$\pm$0.31 & 54.11$\pm$0.32 & 26.56$\pm$0.40 \\
    FedMF & 38.54$\pm$1.87 & 23.74$\pm$0.19 & 45.57$\pm$2.17 & 26.02$\pm$0.24 & 30.72$\pm$0.52 & 16.40$\pm$0.29 & 49.64$\pm$0.38 & 22.51$\pm$0.17 \\
    PerFedRec & 45.82$\pm$2.76 & 33.90$\pm$2.63 & 59.54$\pm$1.55 & 40.07$\pm$0.98 & 36.42$\pm$0.37 & 23.09$\pm$0.27 & 55.02$\pm$0.73 & 29.19$\pm$0.34 \\
    PFedRec & 46.73$\pm$0.32 & 34.04$\pm$0.16 & 59.59$\pm$0.28 & 38.13$\pm$0.06 & \cellcolor[HTML]{C8EBF6} 41.65$\pm$0.34 & \cellcolor[HTML]{C8EBF6} 27.51$\pm$0.24 & \cellcolor[HTML]{C8EBF6} 58.87$\pm$0.34 & \cellcolor[HTML]{C8EBF6} 32.86$\pm$0.26 \\
    FedRAP & 31.90$\pm$1.12 & 25.16$\pm$0.75 & 43.88$\pm$1.14 & 31.47$\pm$0.93 & 20.68$\pm$0.72 & 14.47$\pm$0.51 & 41.57$\pm$0.53 & 24.38$\pm$0.36 \\
    CoFedRec & 48.24$\pm$0.29 & \cellcolor[HTML]{C8EBF6} 35.11$\pm$0.29 & 60.82$\pm$0.20 & \cellcolor[HTML]{C8EBF6} 40.50$\pm$0.20 & 35.68$\pm$0.36 & 21.96$\pm$0.21 & 56.02$\pm$1.06 & 28.06$\pm$0.67 \\
    GPFedRec & 45.66$\pm$1.37 & 32.96$\pm$1.06 & 57.26$\pm$2.17 & 36.74$\pm$1.19 & 34.53$\pm$1.23 & 21.38$\pm$1.42 & 55.20$\pm$0.27 & 27.97$\pm$0.92 \\
    \method{} & \cellcolor[HTML]{95D4EE} \textbf{51.20$\pm$0.02} & \cellcolor[HTML]{95D4EE} \textbf{37.60$\pm$0.11} & \cellcolor[HTML]{95D4EE} \textbf{62.70$\pm$0.14} & \cellcolor[HTML]{95D4EE} \textbf{41.33$\pm$0.16} & \cellcolor[HTML]{95D4EE} \textbf{77.27$\pm$0.09} & \cellcolor[HTML]{95D4EE} \textbf{58.39$\pm$0.21} & \cellcolor[HTML]{95D4EE} \textbf{89.90$\pm$0.03} & \cellcolor[HTML]{95D4EE} \textbf{62.52$\pm$0.19} \\
    \bottomrule
\end{tabular}
\end{table*}

\paragraph{Hyperparameters}
We refer to the open-sourced code provided by these methods, adopting the recommended hyperparameters, and tuning them accordingly.
The local epoch is set to 1, and the total communication round is set to 100.
We set both the user and item embedding sizes to 32, with the batch size fixed at 256.
We adopt the Adam optimizer.
For \method{}, we carry out a grid search on $\rho_{\mathrm{ur}}$ and $\rho_{\mathrm{co}}$ and jointly tune the learning rate, choosing the best hyperparameters separately for each dataset.

\paragraph{Hardware Information}
All models and algorithms are implemented using Python 3.9 and PyTorch 2.3.0.
The experiments are conducted on a server running Ubuntu 22.04, equipped with 256GB of RAM and an NVIDIA GeForce RTX 4090 GPU.

\begin{figure*}[t]
  \centering
  \subfigure[\textbf{FedNCF, FilmTrust dataset}]{\includegraphics[width=0.245\linewidth]{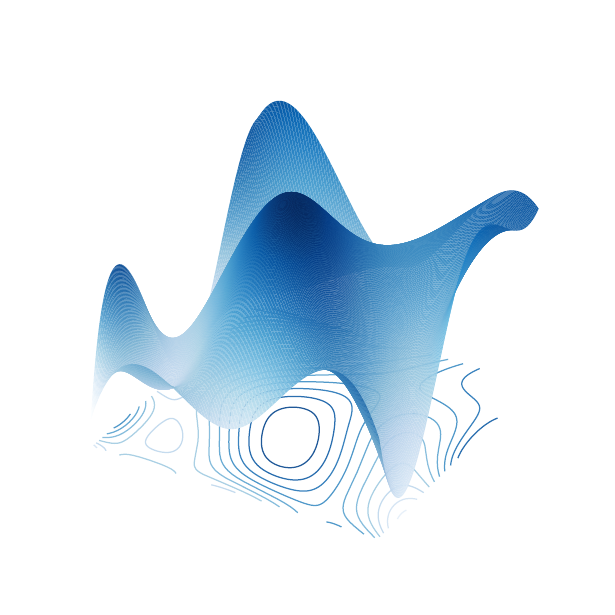}} \label{fig:lt-fedncf-filmtrust}
  \subfigure[\textbf{\method{}, FilmTrust dataset}]{\includegraphics[width=0.245\linewidth]{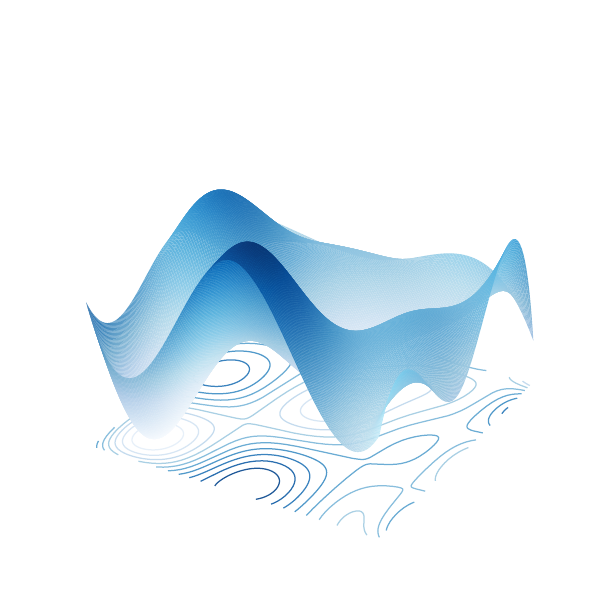}} \label{fig:lt-fedrecgel-filmtrust}
  \subfigure[\textbf{FedNCF, Lastfm-2K dataset}]{\includegraphics[width=0.245\linewidth]{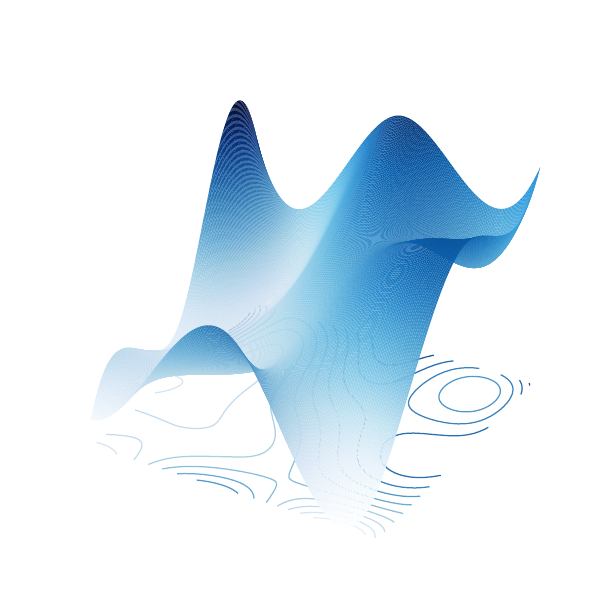}} \label{fig:lt-fedncf-lastfm_2k}
  \subfigure[\textbf{\method{}, Lastfm-2K dataset}]{\includegraphics[width=0.245\linewidth]{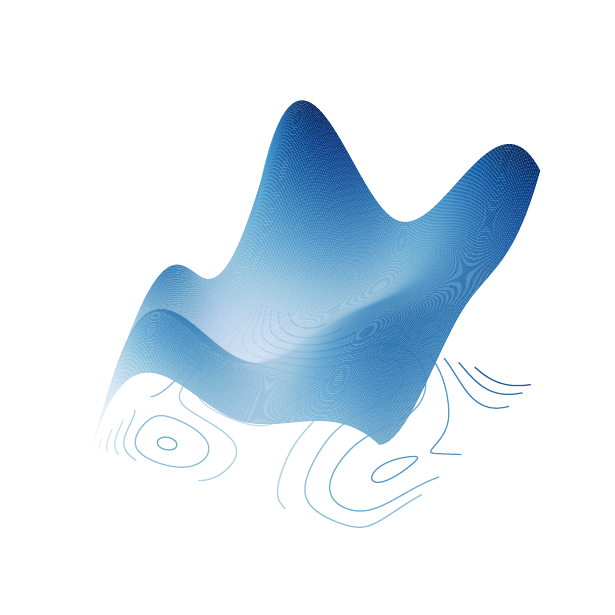}} \label{fig:lt-fedrecgel-lastfm_2k} \\
  \vspace{-1.0em}
  \subfigure[\textbf{FedNCF, Amazon-Video dataset}]{\includegraphics[width=0.245\linewidth]{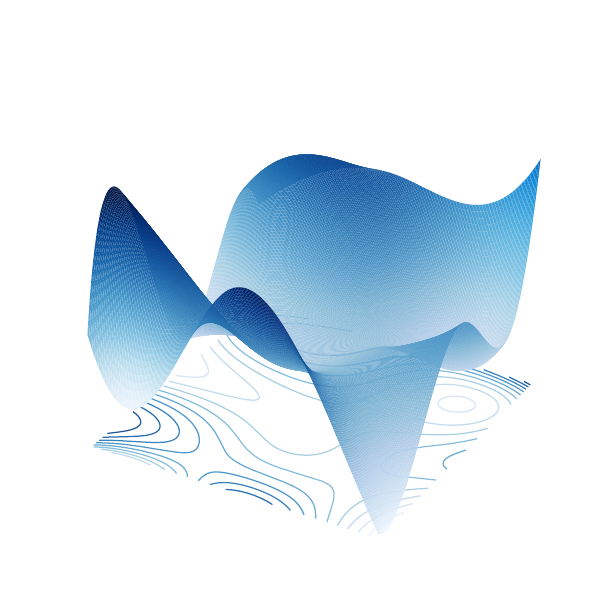}} \label{fig:lt-fedncf-amazon-video}
  \subfigure[\textbf{\method{}, Amazon-Video dataset}]{\includegraphics[width=0.245\linewidth]{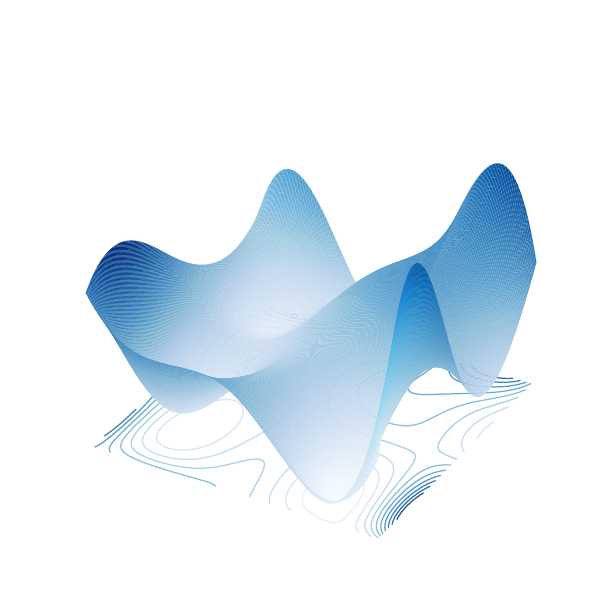}} \label{fig:lt-fedrecgel-amazon-video}
  \subfigure[\textbf{FedNCF, QB-article dataset}]{\includegraphics[width=0.245\linewidth]{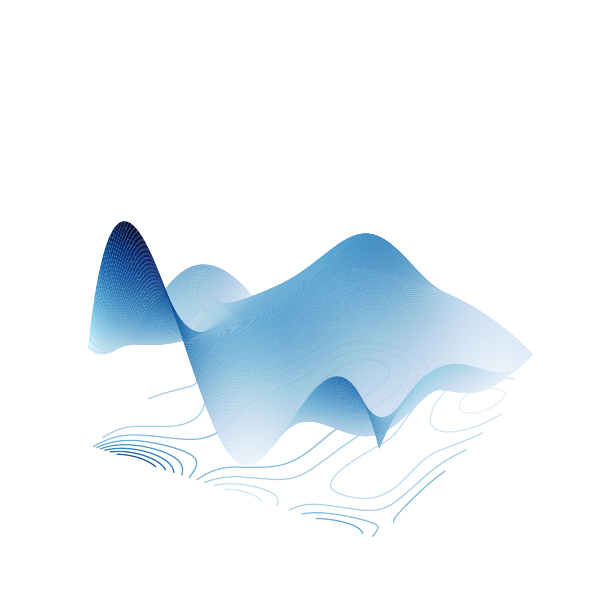}} \label{fig:lt-fedncf-qb_article}
  \subfigure[\textbf{\method{}, QB-article dataset}]{\includegraphics[width=0.245\linewidth]{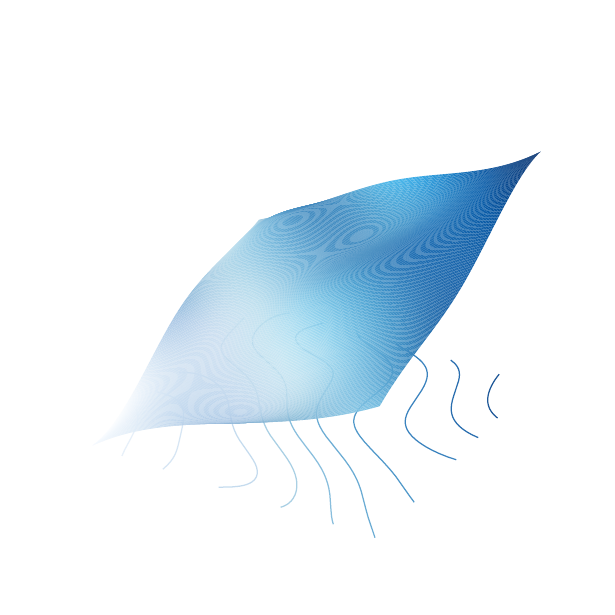}} \label{fig:lt-fedrecgel-qb_article}
  \caption{3D surface plots of the post-convergence landscapes for models trained with \method{} and with FedNCF.}
  \label{fig:loss_landscape}
\end{figure*}

\begin{figure*}[h]
    \centering
    \includegraphics[width=0.3\linewidth]{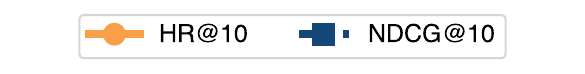} \\
    \vspace{-1.0em}
    \subfigure[$\rho_{\mathrm{ur}}$, \textbf{FilmTrust dataset}]{\includegraphics[width=0.245\linewidth]{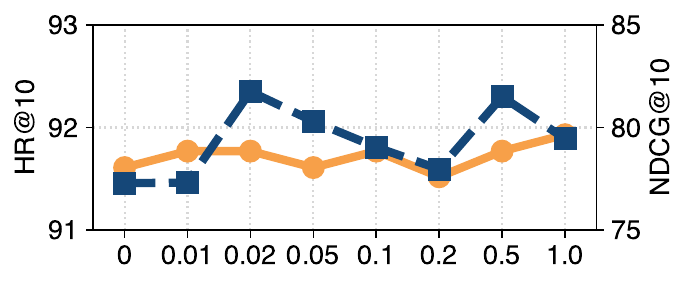}} \label{fig:hp-filmtrust-ns}
    \subfigure[$\rho_{\mathrm{ur}}$, \textbf{Lastfm-2K dataset}]{\includegraphics[width=0.245\linewidth]{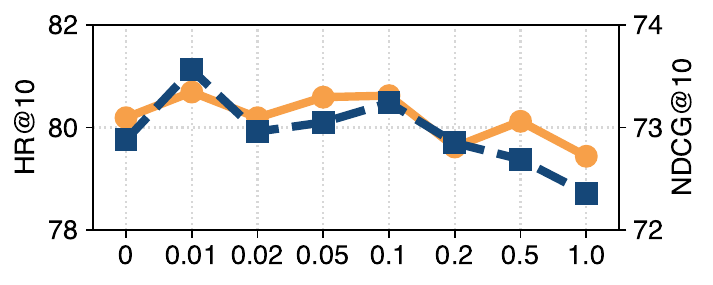}} \label{fig:hp-lastfm-ns}
    \subfigure[$\rho_{\mathrm{ur}}$, \textbf{Amazon-Video dataset}]{\includegraphics[width=0.245\linewidth]{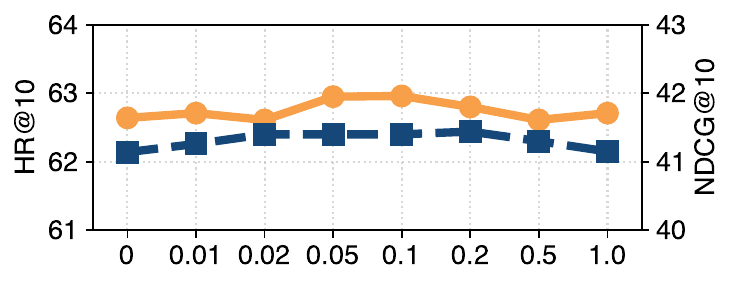}} \label{fig:hp-amazon-ns}
    \subfigure[$\rho_{\mathrm{ur}}$, \textbf{QB-article dataset}]{\includegraphics[width=0.245\linewidth]{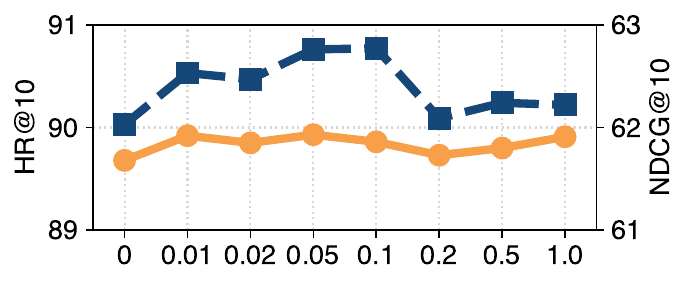}} \label{fig:hp-qb-ns} \\
    \vspace{-1.0em}
    \subfigure[$\rho_{\mathrm{co}}$, \textbf{FilmTrust dataset}]{\includegraphics[width=0.25\linewidth]{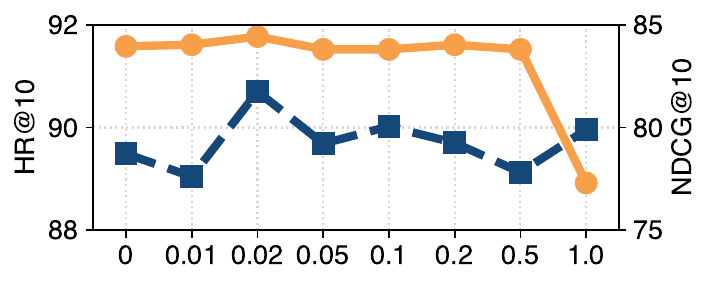}} \label{fig:hp-filmtrust-sh}
    \subfigure[$\rho_{\mathrm{co}}$, \textbf{Lastfm-2K dataset}]{\includegraphics[width=0.24\linewidth]{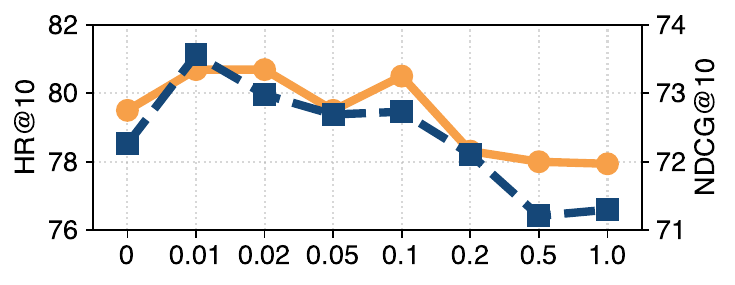}} \label{fig:hp-lastfm-sh}
    \subfigure[$\rho_{\mathrm{co}}$, \textbf{Amazon-Video dataset}]{\includegraphics[width=0.24\linewidth]{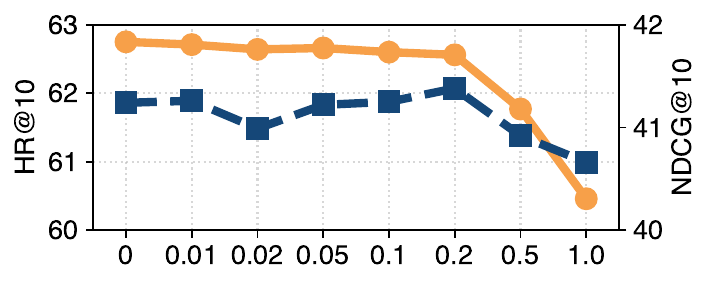}} \label{fig:hp-amazon-sh}
    \subfigure[$\rho_{\mathrm{co}}$, \textbf{QB-article dataset}]{\includegraphics[width=0.245\linewidth]{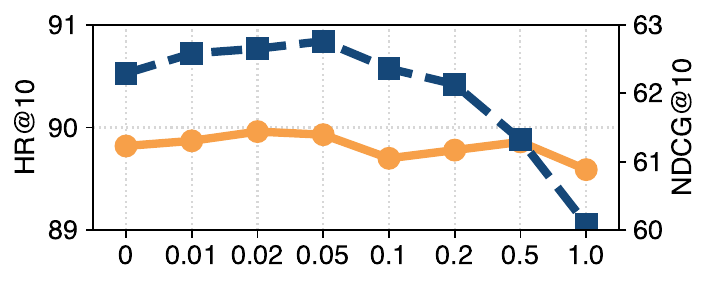}} \label{fig:hp-qb-sh}
    \caption{Effect of hyperparameter $\rho_{\mathrm{ur}}$ and $\rho_{\mathrm{co}}$. We conduct experiments on all four datasets. We report HR@10 and NDCG@10 to represent the recommendation performance.}
    \label{fig:hyperparam}
\end{figure*}

\subsection{Results and Discussions}
\subsubsection{Main Results (RQ1)}
Table~\ref{tab:main_results} shows the average performance results and standard deviation obtained from five runs with different seeds.
We have the following observations and insights:
\begin{itemize}[leftmargin=*]\setlength{\itemsep}{-\itemsep}
    \item Simply combining centralized recommendation methods with federated learning, where model parameters are shared directly, does not lead to optimal results.
    For example, FedNCF and FedMF perform poorly in most settings, because they only minimize the empirical loss and overlook the generalization ability of item embeddings and model parameters.

    \item Methods based on clustering similar users show inconsistent performance across datasets.
    For example, CoFedRec achieves strong performance on Lastfm-2K, but underperforms unexpectedly on FilmTrust.
    This discrepancy is likely due to its heavy reliance on clustering outcomes, which may be unreliable in federated settings because of client sparsity and insufficient user representations.

    \item Methods that train personalized models for each client, such as PFedRec, achieve the best results on Lastfm-2K.
    This is because Lastfm-2K has the lowest user-item ratio, giving each clients abundant local interaction data to learn a personalized model.
    However, in dataset like Amazon-Video and QB-article, these methods perform less well due to insufficient local data for each client.

    \item Our method, \method{}, consistently outperforms all baselines across diverse scenarios.
    Interestingly, its advantage grows as the user-item ratio increases.
    In the dataset with the largest user-item ratio, QB-article, the improvement in HR@10 exceeds 50\%, confirming the effectiveness of \method{} with local SAM training.
    In real world settings, where user-item ratios tend to be much higher, this suggest strong practical viability.
    Even on the dataset with the smallest user-item ratio (Lastfm-2K), \method{} remains comparable to the best baseline (PFedRec).
\end{itemize}

\begin{table*}[h]
\centering
\caption{Ablation study on non-shared and shared parts in SAM training. We compare (i) \method{}-w/o-\emph{non-shared}, which disables SAM on non-shared parts (user embeddings), keeping it on shared parts; (ii) \method{}-w/o-\emph{shared}, which disables SAM on shared parts (item embeddings and parameters), keeping it on non-shared parts; and (iii) \method{}, the full model with SAM applied to both. Results are expressed as percentages (\%).}
\label{tab:ablation}
\begin{tabular}{ccccccccc}
    \toprule
    Dataset & \multicolumn{4}{c}{FilmTrust} & \multicolumn{4}{c}{Lastfm-2K} \\ \cmidrule(lr){2-5} \cmidrule(lr){6-9}
    Methods & HR@5 & NDCG@5 & HR@10 & NDCG@10 & HR@5 & NDCG@5 & HR@10 & NDCG@10 \\ \midrule
    \method{}-w/o-\emph{non-share} & 89.02$\pm$0.14 & 79.54$\pm$2.17 & 91.63$\pm$0.23 & 77.29$\pm$2.20 & 71.88$\pm$0.21 & 69.88$\pm$0.34 & 70.19$\pm$0.37 & 72.88$\pm$2.18 \\
    \method{}-w/o-\emph{shared} & 89.05$\pm$0.17 & 81.23$\pm$0.37 & 91.58$\pm$0.20 & 81.72$\pm$0.86 & 70.69$\pm$0.31 & 69.09$\pm$0.22 & 79.50$\pm$0.12 & 72.26$\pm$0.28 \\
    \method{} & 89.38$\pm$0.21 & 81.33$\pm$0.32 & 91.66$\pm$0.10 & 82.01$\pm$0.23 & 73.92$\pm$0.22 & 70.79$\pm$0.16 & 80.69$\pm$0.25 & 73.56$\pm$0.14 \\
    \bottomrule
    Dataset & \multicolumn{4}{c}{Amazon-Video} & \multicolumn{4}{c}{QB-article} \\ \cmidrule(lr){2-5} \cmidrule(lr){6-9}
    Methods & HR@5 & NDCG@5 & HR@10 & NDCG@10 & HR@5 & NDCG@5 & HR@10 & NDCG@10 \\ \midrule
    \method{}-w/o-\emph{non-shared} & 51.12$\pm$0.01 & 37.39$\pm$0.06 & 62.64$\pm$0.08 & 41.14$\pm$0.10 & 76.92$\pm$0.16 & 57.76$\pm$0.08 & 89.68$\pm$0.05 & 62.03$\pm$0.05 \\
    \method{}-w/o-\emph{shared} & 51.04$\pm$0.27 & 37.34$\pm$0.29 & 61.75$\pm$0.12 & 41.14$\pm$0.22 & 75.35$\pm$0.02 & 56.33$\pm$0.21 & 87.95$\pm$0.06 & 60.45$\pm$0.21 \\
    \method{} & 51.20$\pm$0.02 & 37.60$\pm$0.11 & 62.70$\pm$0.14 & 41.33$\pm$0.16 & 77.27$\pm$0.09 & 58.39$\pm$0.21 & 89.90$\pm$0.03 & 62.52$\pm$0.19 \\
    \bottomrule
\end{tabular}
\end{table*}

\subsubsection{Hyperparameter Sensitivity (RQ2)}
To assess hyperparameter sensitivity, we run experiments on all four datasets.
\method{} has two key hyperparameters: $\rho_{\mathrm{ur}}$, the neighborhood radius used to search the worst-case perturbation for the non-shared parts (user embeddings), and $\rho_{\mathrm{co}}$, the corresponding radius for the shared parts (item embeddings and model parameters).
When varying one hyperparameter, we fix the other to its dataset-specific optimal value.
For each dataset, we sweep $\rho_{\mathrm{ur}}$ and $\rho_{\mathrm{co}}$ over the grid $[0,0.01,0.02,0.05,0.1,0.2,0.5,1.0]$.
The results are shown in Figure~\ref{fig:hyperparam}.
We have the following observations and insights:
\begin{itemize}[leftmargin=*]\setlength{\itemsep}{-\itemsep}
    \item Compare to HR, NDCG exhibits greater sensitivity to both hyperparameters.
    This is attribute to NDCG being a position-aware ranking metric, which incorporates more detailed information than HR.
    
    \item Across all datasets, HR remains unaffected by variations in $\rho_{\textrm{ns}}$.
    In contrast, HR remains constant between 0 and 0.2 for $\rho_{\textrm{sh}}$, but experiences a significant decline from 0.2 to 1.0, particularly in the FilmTrust and Amazon-Video datasets.
    This decline is due to the fact that when the neighborhood size is excessively large, the computed worst-case perturbation fails to serve as a solution for the inner maximization in problem~\eqref{eq:sam}, owing to the properties of Taylor expansion.

    \item For smaller datasets, NDCG exhibits fluctuations without a clear pattern, underscoring the importance of selecting appropriate hyperparameters in such context.
    However, empirical results indicate that the optimal hyperparameter typically falls within the range of 0.02 to 0.2, thereby reducing the complexity of the hyperparameter search process.
    
    \item For larger datasets, such as Amazon-Video and QB-article, NDCG follows a bell-shaped curve, increasing from 0 to approximately 0.1, then decreasing from 0.1 to 1.0.
    This pattern arises because, with a smaller neighborhood size, the perturbation cannot reach a worst-case perturbation, diminishing the effectiveness of SAM training.
    Consequently, as the neighborhood size increases, NDCG improves.
    However, as the neighborhood size becomes too large, similar to the reason for the decline in HR, the computed worst-case perturbation no longer serves as a solution for the inner maximization in problem~\eqref{eq:sam}, due to the properties of Taylor expansion.
\end{itemize}

\subsubsection{Ablation Study (RQ3)}
We conduct an ablation study across all four datasets to verify the contributions of SAM training on shared and non-shared parts.
Specifically, We compare:
\begin{itemize}[leftmargin=*]\setlength{\itemsep}{-\itemsep}
    \item \method{}-w/o-\emph{non-shared}: SAM is disabled on the non-shared parts (i.e., user embeddings), while still applied on the shared parts;
    
    \item \method{}-w/o-\emph{shared}: SAM is disabled on the shared parts (item embeddings and model parameters), but still applied on the non-shared parts;
    
    \item \method{}, the full method, with SAM on both shared and non-shared parts.
\end{itemize}
The results demonstrate that \method{} outperforms \method{}-w/o-non-shared, which in turn outperforms \method{}-w/o-shared.
This indicates that both components contribute, but SAM on shared parts plays a more critical role in the federated training process, enabling cross-client information exchange, whereas SAM on the non-shared parts primarily enhances stability at each client.

\subsubsection{Loss Landscape (RQ4)}
To explore the role of SAM training in \method{}, we plot the loss landscapes.
We randomly perturb two dimensions of the model parameters and compute the corresponding loss.
Figure~\ref{fig:loss_landscape} presents 3D surface visualizations of the post-convergence loss landscapes for models trained with \method{} and FedNCF across all four datasets.
Prior work suggests that flatter loss surfaces correlate with better generalization.
As seen in Figure~\ref{fig:loss_landscape}~(a) and \ref{fig:loss_landscape}~(b), the model trained via FedNCF does not settle in a valley, whereas \method{} lands in a relatively flat basin.
Similarly, in Figures \ref{fig:loss_landscape}~(c) and \ref{fig:loss_landscape}~(d), \method{} yields a flatter surrounding landscape, while FedNCF’s converged point lies on a ridge.
These visualizations suggest that incorporating SAM during local training smooths the loss surface, leading to item embeddings and parameters with stronger generalization, which in turn improves downstream recommendation performance. 
We include contour map visualizations of the loss landscapes in Appendix \ref{sec:loss_landscape}. Since we do not annotate scales in the main figure, we also provide average loss values at different perturbation magnitudes in the appendix.

%% file: body/06_conclusion.tex
\section{Conclusion}
In this paper, we studied federated recommender systems with the objective of learning generalized embeddings to enhance recommendation performance.
Existing methods overlook a \textbf{critical issue}, i.e., the stable learning of a generalized item embedding throughout the federated recommender system training process.
Item embedding plays a central role in facilitating knowledge sharing across clients.
Yet, under the cross-device setting, local data distributions exhibit significant heterogeneity and sparsity, exacerbating the difficulty of learning generalized embeddings.
These factors make the stable learning of generalized item embeddings both indispensable for effective federated recommendation and inherently difficult to achieve.

To fill this gap, we propose a new federated recommendation framework, named \textbf{Fed}erated \textbf{Rec}ommendation with \textbf{G}eneralized \textbf{E}mbedding \textbf{L}earning (\method{}).
We reformulate the federated recommendation problem from an item-centered perspective and cast it as a multi-task learning problem, aiming to learn generalized embeddings throughout the training procedure.
Based on theoretical analysis, we employ sharpness-aware minimization to address the generalization problem, thereby stabilizing the training process and enhancing recommendation performance.

We conduct extensive experiments on four real-world datasets to evaluate the effectiveness of our proposed method.
Our method, \method{}, consistently outperforms all baselines across diverse scenarios.
Its advantage grows as the user-item ratio increases.
Due to considerations of computational efficiency, we adopted a simple aggregation scheme in the global aggregation phase.
As future directions, it would be interesting to investigate alternative aggregation strategies and to extend our framework to more diverse federated recommendation scenarios and methodologies.

%% file: body/appendix.tex
\section{Detailed Derivation of Sharpness-Aware Minimization}\label{sec:sam_solution}
The optimization problem~\eqref{eq:sam} is referred to as shapness-aware minimization (SAM).
We first derive the solution of the inner maximization problem, the worst-case perturbation $\boldsymbol{\epsilon}^*$, then compute the gradient $\mathbf{g}^{\mathrm{SAM}}$ with respect to the peturbed model to optimize the outer minimization problem. 

Assume that the loss function $\mathcal{L}$ is sufficiently smooth with respect to the parameter $\boldsymbol{\theta}$ (at least locally differentiable). Then, we can perform a first-order Taylor expansion around $\boldsymbol{\theta}$:
\begin{equation*}
\mathcal{L}(\boldsymbol{\theta} + \boldsymbol{\epsilon}) \approx \mathcal{L}(\boldsymbol{\theta}) + \boldsymbol{\epsilon}^\top \nabla_{\boldsymbol{\theta}} \mathcal{L}(\boldsymbol{\theta}) + \mathcal{O}(\|\boldsymbol{\epsilon}\|^2).
\end{equation*}
Since we are interested in small perturbations such that $\|\boldsymbol{\epsilon}\| \leq \rho$ and $\rho$ is small, the higher-order term $\mathcal{O}(\|\boldsymbol{\epsilon}\|^2)$ can be ignored. Thus, the inner maximization becomes
\begin{equation*}
\max_{\|\boldsymbol{\epsilon}\| \leq \rho} \mathcal{L}(\boldsymbol{\theta} + \boldsymbol{\epsilon}) \approx \max_{\|\boldsymbol{\epsilon}\| \leq \rho} \left\{ \mathcal{L}(\boldsymbol{\theta}) + \boldsymbol{\epsilon}^\top \nabla_{\boldsymbol{\theta}} \mathcal{L}(\boldsymbol{\theta}) \right\}.
\end{equation*}
Note that $\mathcal{L}(\boldsymbol{\theta})$ is independent of $\boldsymbol{\epsilon}$, so it can be treated as a constant and removed from the optimization. The remaining subproblem is
\begin{equation*}
\arg\max_{\|\boldsymbol{\epsilon}\| \leq \rho} \boldsymbol{\epsilon}^\top \nabla_{\boldsymbol{\theta}} \mathcal{L}(\boldsymbol{\theta}).
\end{equation*}
We now aim to solve this subproblem:
\begin{equation*}
\max_{\|\boldsymbol{\epsilon}\|_2 \leq \rho} \boldsymbol{\epsilon}^\top \mathbf{g}, \quad \text{where } \mathbf{g} := \nabla_{\boldsymbol{\theta}} \mathcal{L}(\boldsymbol{\theta}).
\end{equation*}
This is a classical linear maximization problem over a Euclidean ball. Geometrically, $\boldsymbol{\epsilon}^\top \mathbf{g}$ is the projection of $\boldsymbol{\epsilon}$ onto the direction of $\mathbf{g}$, with magnitude proportional to $\|\mathbf{g}\|$. 
To maximize the projection, $\boldsymbol{\epsilon}$ should lie on the surface of the $\rho$-ball in the direction of $\mathbf{g}$ (Lagrange multiplier method). 
Hence, the optimal solution is
\begin{equation*}
\boldsymbol{\epsilon}^* = \rho \frac{\mathbf{g}}{\|\mathbf{g}\|_2} = \rho \frac{\nabla_{\boldsymbol{\theta}} \mathcal{L}(\boldsymbol{\theta})}{\|\nabla_{\boldsymbol{\theta}} \mathcal{L}(\boldsymbol{\theta})\|_2}.
\end{equation*}
This $\boldsymbol{\epsilon}^*$ represents the perturbation that most adversely affects the model by pointing in the worst-case direction. With this $\boldsymbol{\epsilon}^*$, the gradient with respect to this perturbed model is computed to update $\boldsymbol{\theta}$:
\begin{equation*}
\mathbf{g}^{\mathrm{SAM}} = \nabla_{\boldsymbol{\theta}} \left[ \max_{\|\boldsymbol{\epsilon}\| \leq \rho} \mathcal{L}(\boldsymbol{\theta} + \boldsymbol{\epsilon}) \right] \approx \nabla_{\boldsymbol{\theta}} \mathcal{L}(\boldsymbol{\theta} + \boldsymbol{\epsilon}^*).
\end{equation*}

\section{Proof of Lemma~\ref{lem:multi-gau-pac-bound}}\label{sec:proof_lem_1}

\begin{proof}
    To prove Lemma~\ref{lem:multi-gau-pac-bound}, we first establish the inequality for the scalar case by fixing $k$ to a specific value.
    This provides the fundamental bound for a single parameter vector $\boldsymbol{\theta}$.
    We then extend the result to the vector form by applying mathematical induction over the index $k \in [m]$.
    This step-by-step strategy allows us to generalize from the single-task case to the multi-task setting considered in the lemma.

    We first establish the inequality for the scalar case.
    We use the PAC-Bayes theory with $\mathcal{P} = \mathcal{N}\bigl(0,\sigma_P^2 \mathbb{I}_T\bigr)$ and $\mathcal{Q} = \mathcal{N}\bigl(\boldsymbol{\theta},\sigma^2 \mathbb{I}_T\bigr)$ as the prior and posterior distributions, respectively.
    By the PAC-Bound~\cite{alquier2016properties}, with probability at least $1-\gamma$ and for all $\beta>0$, we have
    \begin{equation*}
        \mathbb{E}_{\boldsymbol{\theta}\sim \mathcal{Q}}\bigl[\mathcal{L}_\mathcal{D}(\boldsymbol{\theta})\bigr] \le \mathbb{E}_{\boldsymbol{\theta}\sim \mathcal{Q}}\bigl[\mathcal{L}_\mathcal{S}(\boldsymbol{\theta})\bigr] + \frac{1}{\beta}\Bigl[\mathrm{KL}(\mathcal{Q}\|\mathcal{P}) +\log\frac{1}{\gamma} +\Psi(\beta,N) \Bigr],
    \end{equation*}
    where 
    \begin{equation*}
        \Psi(\beta,N)=\log \mathbb{E}_{\mathcal{P}} \mathbb{E}_{\mathcal{S}}\Bigl[\exp\{\beta(\mathcal{L}_{\mathcal{D}}(\boldsymbol{\theta})-\mathcal{L}_{\mathcal{S}}(\boldsymbol{\theta}))\}\Bigr].
    \end{equation*}
    Since the loss is bounded by $L$, Hoeffding's lemma gives
    \begin{equation*}
        \Psi(\beta,N) \le \frac{\beta^2L^2}{8N}.
    \end{equation*}
    By the Cauchy–Schwarz inequality,
    \begin{equation*}
        \frac{1}{\sqrt{N}} \left[\frac{T}{2}\log\!\left(1+\frac{\|\boldsymbol{\theta}\|^2}{T\sigma^2}\right)+\frac{L^2}{8}\right] \ge \frac{L}{2\sqrt{N}}\sqrt{ T\log\!\left(1+\frac{\|\boldsymbol{\theta}\|^2}{T\sigma^2}\right) } \ge L,
    \end{equation*}
    which completes the proof under the assumption that the loss is bounded by $L$.
    Now it remains to prove the theorem in the case
    \begin{equation*}
    \|\boldsymbol{\theta}\|^2  \le  T\sigma^2\bigl[\exp(4N/T)-1\bigr].
    \end{equation*}
    Since the prior $\mathcal{P}$ must be chosen in advance, but its optimal variance depends on $\boldsymbol{\theta}$, we construct a family of priors
    \begin{align*}
        \mathfrak{P} = \Bigl\{ 
            \mathcal{P}_j = \mathcal{N}\bigl(0,\sigma_{\mathcal{P}_j}^2 \mathbb{I}_T \bigr): 
            \ & \sigma_{\mathcal{P}_j}^2 = c \cdot \exp \!\left(\dfrac{1-j}{T} \right), \\
            & c = \sigma^2 \bigl(1+\exp(4N/T)\bigr), \ j=1,2,\dots 
        \Bigr\}.
    \end{align*}
    Set $\gamma_j = \frac{6\gamma}{\pi^2 j^2}$, and the below inequality holds with probability at least $1-\gamma_j$:
    \begin{equation*}
        \mathbb{E}_{\boldsymbol{\theta}\sim \mathcal{Q}} \bigl[\mathcal{L}_{\mathcal{D}}(\boldsymbol{\theta}) \bigr] \le \mathbb{E}_{\boldsymbol{\theta}\sim \mathcal{Q}} \bigl[\mathcal{L}_{\mathcal{S}} (\boldsymbol{\theta})\bigr] + \frac{1}{\beta}\Bigl[\mathrm{KL}(\mathcal{Q}\|\mathcal{P}_j)+\log\frac{1}{\gamma_j}+\frac{\beta^2 L^2}{8N}\Bigr].
    \end{equation*}
    Or it can be written as:
    \begin{align*}
        \mathbb{E}_{\boldsymbol{\varepsilon} \sim \mathcal{N}(0,\sigma^2 \mathbb{I}_T)} 
        \bigl[\mathcal{L}_{\mathcal{D}} (\boldsymbol{\theta}+\boldsymbol{\varepsilon}) \bigr]
        &\le 
        \mathbb{E}_{\boldsymbol{\varepsilon} \sim \mathcal{N}(0,\sigma^2 \mathbb{I}_T)} 
        \bigl[\mathcal{L}_{\mathcal{S}} (\boldsymbol{\theta}+\boldsymbol{\varepsilon})\bigr] \\
        &\quad + \frac{1}{\beta}\Bigl[\mathrm{KL}(\mathcal{Q}\|\mathcal{P}_j)
        + \log\frac{1}{\gamma_j} + \frac{\beta^2 L^2}{8N}\Bigr].
    \end{align*}
    Thus, with probability $1-\gamma$, the above inequalities hold for all $\mathcal{P}_j$. 
    We choose:
    \begin{equation*}
    j^*  = \left\lfloor 1 + T\log\! \left(\dfrac{\sigma^2(1+\exp(4N/T))}{\sigma^2 + \|\boldsymbol{\theta}\|^2/T} \right)\right\rfloor.
    \end{equation*}
    Since $\|\boldsymbol{\theta}\|^2/T \le \sigma^2[\exp(4N/T)-1] $, we get:
    \begin{equation*}
        \sigma^2 + \|\boldsymbol{\theta}\|^2/T  \le  \sigma^2 \exp(4N/T), 
    \end{equation*}
    thus $j^*$ is well-defined. 
    We also have:
    \begin{align*}
        & T\log\frac{c}{\sigma^2 + \|\boldsymbol{\theta}\|^2/T} \le  j^* \le  1 + T\log\frac{c}{\sigma^2 + \|\boldsymbol{\theta}\|^2/T} \\
        \implies \quad & \sigma^2 + \|\boldsymbol{\theta}\|^2/T \le \sigma^2_{P_{j^*}} \le  e^{1/T} (\sigma^2 + \|\boldsymbol{\theta}\|^2/T).
    \end{align*}
    Hence, we have:
    \begin{align*}
        \mathrm{KL}(\mathcal{Q}\|\mathcal{P}_{j^*})
        &=\dfrac12\left[\frac{T\sigma^2+\|\boldsymbol{\theta}\|^2}{\sigma^2_{P_{j^*}}}
        - T + T\log\frac{\sigma^2_{\mathcal{P}_{j^*}}}{\sigma^2}\right]
        \\
        &\le \dfrac12\left[1 + T\log \left(1 + \dfrac{\|\boldsymbol{\theta}\|^2}{T\sigma^2}\right)\right].
    \end{align*}
    For the term $\log\frac{1}{\gamma_{j^*}}$, use the inequality $\log(1+e^t)\le1+t$ for $t>0$:
    \begin{align*}
        \log\frac{1}{\gamma_{j^*}}
        & = \log\frac{(j^*)^2\pi^2}{6\gamma} = \log\frac{1}{\gamma}+\log\frac{\pi^2}{6}+2\log(j^*) \\
        &\le\log\frac{1}{\gamma}+\log\frac{\pi^2}{6} + 2\log\left(1+T\left(1+4N/T\right)\right) \\
        &\le\log\frac{1}{\gamma}+\log\frac{\pi^2}{6} + \log\bigl(1+T+4N \bigr).
    \end{align*}
    Choosing $\beta=\sqrt{N}$, with probability at least $1-\gamma$ we get:
    \begin{align*}
        &\frac{1}{\beta}\Bigl[\mathrm{KL}(\mathcal{Q}\|\mathcal{P}_{j^*})
          + \log\frac{1}{\gamma_{j^*}} + \frac{\beta^2L^2}{8N}\Bigr]\\
        &\le \frac{1}{\sqrt{N}}\left[\dfrac{1}{2}+\dfrac{T}{2} 
          \log \left(1+\dfrac{\|\boldsymbol{\theta}\|^2}{T\sigma^2}\right)
          + \log\dfrac{1}{\gamma} + 6\log(N+T)\right] + \frac{L^2}{8\sqrt{N}}.
    \end{align*}
    The scalar case is proved.

    We extend the result to vector form by applying mathematical induction.
    The result for the base case $k=1$ follows by applying the result of the scalar case with  $\xi=\gamma$ and defining $f^1$ accordingly. 
    
    Now assume Lemma~\ref{lem:multi-gau-pac-bound} holds for all $k\in[n]$ with probability $1-\gamma/2$; that is,
    \begin{equation*}
        \bigl[\mathcal{L}_{\mathcal{D}}(\boldsymbol{\theta}^k)\bigr]_{k=1}^n \le \bigl[\mathbb{E}_{\boldsymbol{\varepsilon} \sim \mathcal{N}(0,\sigma^2 \mathbb{I})} \mathcal{L}_{\mathcal{S}}(\boldsymbol{\theta}^k + \boldsymbol{\varepsilon}) +f^k(\|\boldsymbol{\theta}^k\|_2^2)\bigr]_{k=1}^n.
    \end{equation*}
    Applying the result of the scalar case to $\boldsymbol{\theta}^{n+1} $ with $\xi=\gamma/2$, we get with probability $1-\gamma/2$:
    \begin{equation*}
        \mathcal{L}_{\mathcal{D}}(\boldsymbol{\theta}^{n+1}) \le \mathbb{E}_{\boldsymbol{\varepsilon} \sim \mathcal{N}(0,\sigma^2 \mathbb{I})}
        \bigl[\mathcal{L}_{\mathcal{S}}(\boldsymbol{\theta}^{n+1}+\boldsymbol{\varepsilon})\bigr] + f^{n+1}(\|\boldsymbol{\theta}^{n+1}\|_2^2).
    \end{equation*}
    By the inclusion–exclusion principle, with overall probability at least $1-\gamma$, the claim holds for $m=n+1$, completing the induction.
\end{proof}

\begin{figure*}[t]
  \centering
  \subfigure[\textbf{FedNCF, FilmTrust dataset}]{\includegraphics[width=0.23\linewidth]{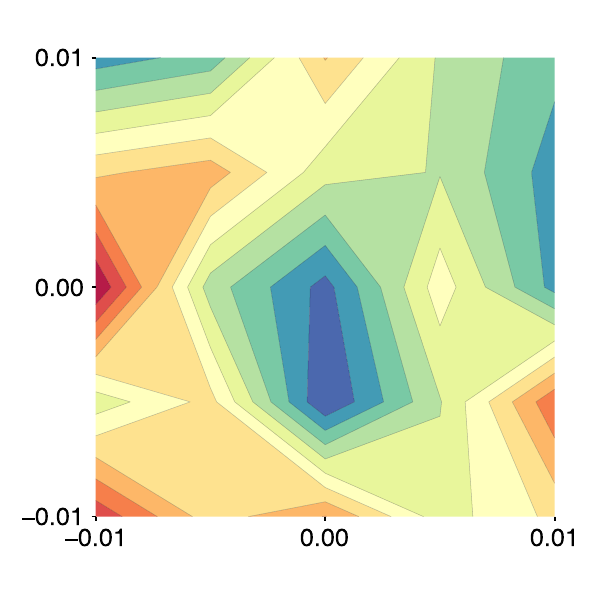}} \label{fig:lc-fedncf-filmtrust}
  \subfigure[\textbf{\textbf{\method{}, FilmTrust dataset}}]{\includegraphics[width=0.24\linewidth]{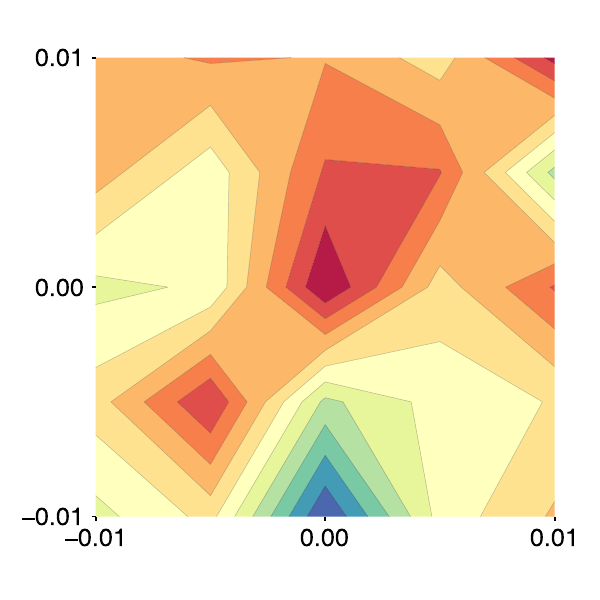}} \label{fig:lc-fedrecgel-filmtrust}
  \subfigure[\textbf{FedNCF, Lastfm-2K}]{\includegraphics[width=0.24\linewidth]{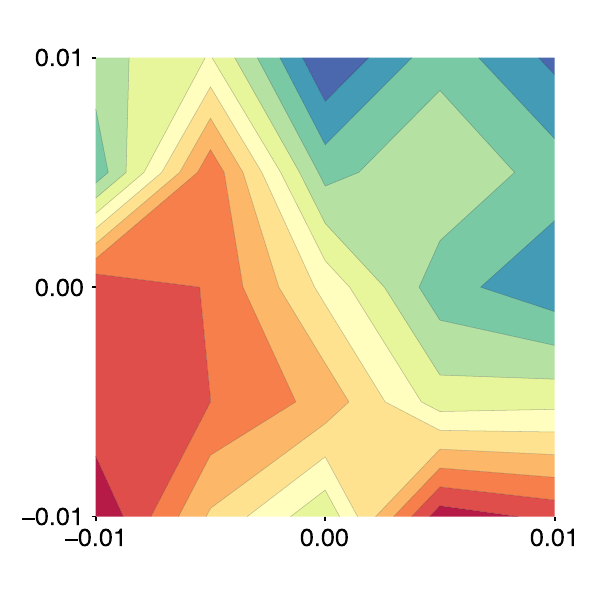}} \label{fig:lc-fedncf-lastfm_2k}
  \subfigure[\textbf{\method{}, Lastfm-2K}]{\includegraphics[width=0.24\linewidth]{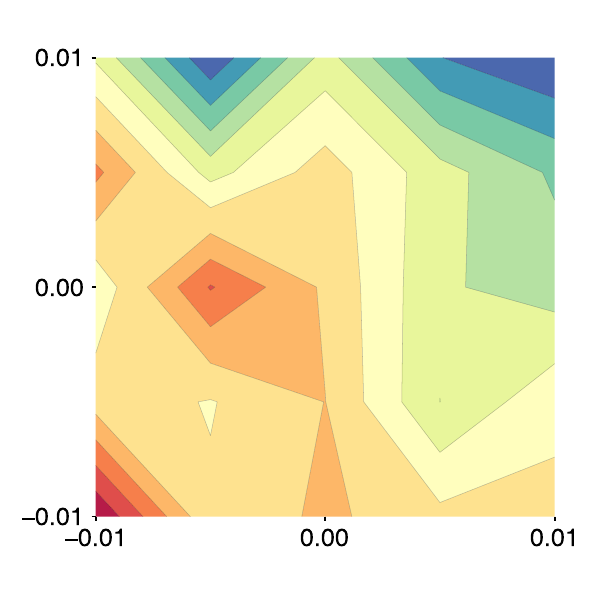}} \label{fig:lc-fedrecgel-lastfm_2k} \\
  \vspace{-1.0em}
  \subfigure[\textbf{FedNCF, Amazon-Video dataset}]{\includegraphics[width=0.24\linewidth]{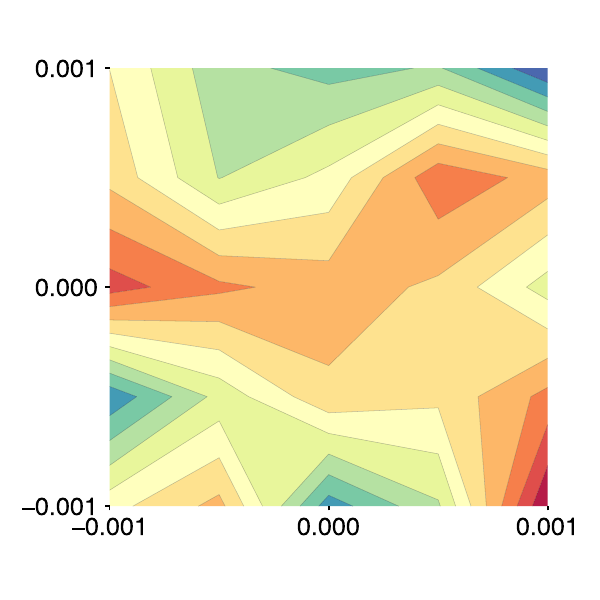}} \label{fig:lc-fedncf-amazon-video}
  \subfigure[\textbf{\textbf{\method{}, Amazon-Video dataset}}]{\includegraphics[width=0.24\linewidth]{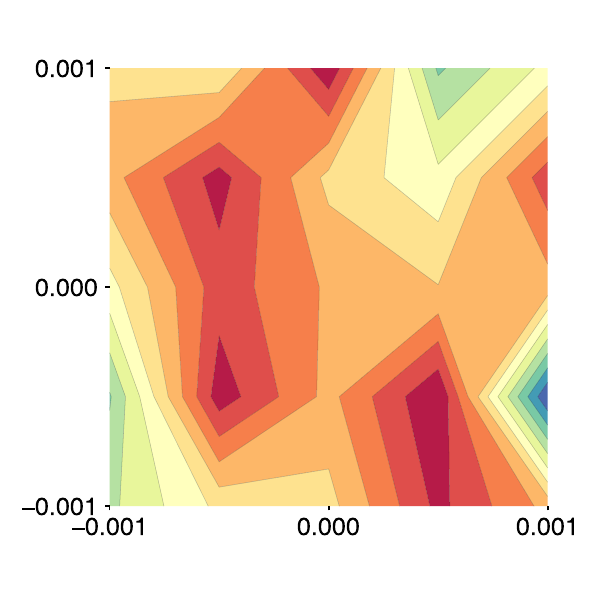}} \label{fig:lc-fedrecgel-amazon-video}
  \subfigure[\textbf{FedNCF, QB-article}]{\includegraphics[width=0.24\linewidth]{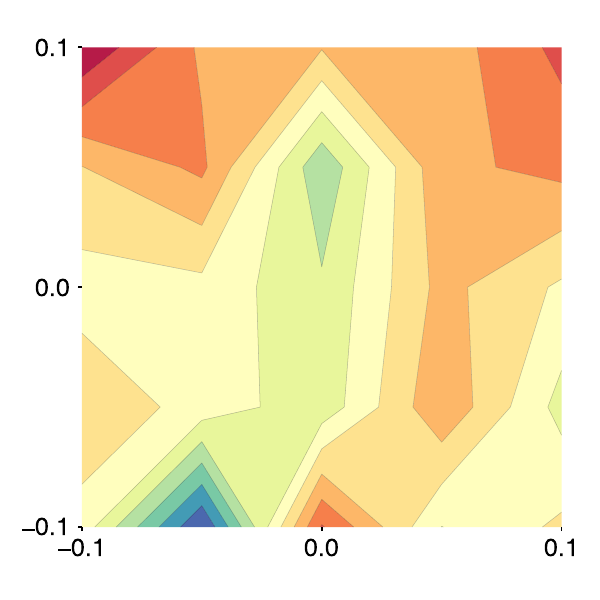}} \label{fig:lc-fedncf-qb_article}
  \subfigure[\textbf{\method{}, QB-article}]{\includegraphics[width=0.24\linewidth]{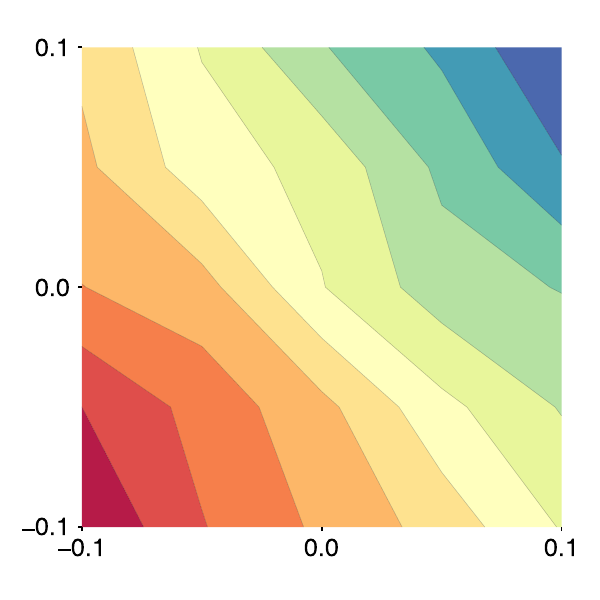}} \label{fig:lc-fedrecgel-qb_article}
  \caption{The contour map visualization of the post-convergence landscapes for models trained with \method{} and with FedNCF across all four datasets.}
  \label{fig:loss_landscape_appendix}
\end{figure*}

\section{Proof of Lemma~\ref{lem:hie-sam}}\label{sec:proof_lem_2}
\begin{proof}
    From Lemma~\ref{lem:multi-gau-pac-bound}, since $\boldsymbol{\theta}^k=(\boldsymbol{\theta}_{\mathrm{co}},\boldsymbol{\theta}_{\mathrm{ur}}^k)$, writing the expectation as an iterated integral we get
    \begin{align*}
        \bigl[\mathcal{L}_{\mathcal{D}}(\boldsymbol{\theta}^k)\bigr]_{k=1}^m 
        & \le \bigl[\mathbb{E}_{\boldsymbol{\varepsilon}\sim\mathcal{N}(0, \sigma^2 \mathbb{I})} \mathcal{L}_{\mathcal{S}}(\boldsymbol{\theta}^k+\boldsymbol{\varepsilon})+f^k(\|\boldsymbol{\theta}^k\|_2^2)\bigr]_{k=1}^m \\
        & = \biggl[\int \mathbb{E}_{\boldsymbol{\varepsilon}_{\mathrm{ur}}^k} \bigl[\mathcal{L}_{\mathcal{S}}(\boldsymbol{\theta}_{\mathrm{co}}+\boldsymbol{\varepsilon}_{\mathrm{co}},\boldsymbol{\theta}_{\mathrm{ur}}^k+\boldsymbol{\varepsilon}_{\mathrm{ur}}^k)\bigr]p(\boldsymbol{\varepsilon}_{\mathrm{co}})d\boldsymbol{\varepsilon}_{\mathrm{co}}\\
        &\quad+f^k(\|\boldsymbol{\theta}^k\|_2^2)\biggr]_{k=1}^m \\
        & = \mathbb{E}_{\boldsymbol{\varepsilon}_{\mathrm{co}}}\Bigl[\mathbb{E}_{\boldsymbol{\varepsilon}_{\mathrm{ur}}^k} \bigl[\mathcal{L}_{\mathcal{S}}(\boldsymbol{\theta}_{\mathrm{co}}+\boldsymbol{\varepsilon}_{\mathrm{co}},\boldsymbol{\theta}_{\mathrm{ur}}^k+\boldsymbol{\varepsilon}_{\mathrm{ur}}^k)\bigr]+f^k(\|\boldsymbol{\theta}^k\|_2^2)\Bigr]_{k=1}^m,
    \end{align*}
    where $p(\boldsymbol{\varepsilon}_{\mathrm{co}})$ is the density of $\mathcal{N}(0,\sigma^2 \mathbb{I}_{T_{\mathrm{co}}})$.
    
    We have $\boldsymbol{\varepsilon}_{\mathrm{ur}}^k\sim\mathcal{N}(0,\sigma^2 \mathbb{I}_{T_{k,\mathrm{ur}}})$ with the dimension $T_{k,\mathrm{ur}}$, so $\|\boldsymbol{\varepsilon}_{\mathrm{ur}}^k\|_2^2$ follows a Chi‐square distribution. 
    As proven in Chi-square distribution bound~\cite{laurent2000adaptive}, for all $k$ and all $t>0$:
    \begin{gather*}
        P\Bigl(\|\boldsymbol{\varepsilon}_{\mathrm{ur}}^k\|_2^2 \ge T_{k,\mathrm{ur}}\sigma^2 + 2 \sigma^2 \sqrt{T_{k,\mathrm{ur}}t} + 2t \sigma^2\Bigr) \le e^{-t},\\
        P\Bigl(\|\boldsymbol{\varepsilon}_{\mathrm{ur}}^k\|_2^2 < T_{k,\mathrm{ur}}\sigma^2 + 2\sigma^2\sqrt{T_{k,\mathrm{ur}}t}+2t\sigma^2\Bigr) > 1-e^{-t}.
    \end{gather*}
    Select $t=\ln(\sqrt{N})$, we derive the following bound for the noise magnitude in terms of the perturbation radius $\rho_{\mathrm{ur}}$ for all $k$:
    \begin{equation}
        P\left(\|\boldsymbol{\varepsilon}_{\mathrm{ur}}^k\|_2^2 \le \sigma^2 \left(2 \ln\left(\sqrt{N}\right) +T_{k,\mathrm{ur}} +2\sqrt{T_{k,\mathrm{ur}} \ln\left(\sqrt{N}\right)} \right) \right) > 1-\frac{1}{\sqrt{N}}.
        \label{eq:lem2:1} \tag{Lem2:1}
    \end{equation}
    
    Moreover, we have $\boldsymbol{\varepsilon}_{\mathrm{co}} \sim \mathcal{N}(0,\sigma^2 \mathbb{I}_{T_{\mathrm{co}}})$ with the dimension $T_{\mathrm{co}}$, so $\|\boldsymbol{\varepsilon}_{\mathrm{co}}\|_2^2$ follows the Chi‐square distribution. 
    From Chi-square distribution bound~\cite{laurent2000adaptive}, for all $t>0$:
    \begin{gather*}
        P\Bigl(\|\boldsymbol{\varepsilon}_{\mathrm{co}}\|_2^2 \ge T_{\mathrm{co}}\sigma^2 + 2\sigma^2\sqrt{T_{\mathrm{co}}t}+2t\sigma^2\Bigr)\le e^{-t},\\
        P\Bigl(\|\boldsymbol{\varepsilon}_{\mathrm{co}}\|_2^2 < T_{\mathrm{co}}\sigma^2 + 2\sigma^2\sqrt{T_{\mathrm{co}}t}+2t\sigma^2\Bigr)>1-e^{-t}.
    \end{gather*}
    Again, take $t=\ln(\sqrt{N})$, we derive the following bound for the noise magnitude in terms of the perturbation radius $\rho_{\mathrm{co}}$
    \begin{equation}
        P\left(\|\boldsymbol{\varepsilon}_{\mathrm{co}}\|_2^2 \le \sigma^2 \left(2 \ln\left(\sqrt{N}\right) + T_{\mathrm{co}} + 2\sqrt{T_{\mathrm{co}} \ln\left(\sqrt{N}\right)} \right) \right) > 1-\frac{1}{\sqrt{N}}.
        \label{eq:lem2:2} \tag{Lem2:2}
    \end{equation}
    
    By choosing
    \begin{align*}
        \sigma < \min\Bigl\{&\frac{\rho_{\mathrm{co}}}{\sqrt{2\ln N^{1/2}+T_{\mathrm{co}}+2\sqrt{T_{\mathrm{co}}\ln N^{1/2}}}}, \\
        & \min_i\frac{\rho_{\mathrm{ur}}}{\sqrt{2\ln N^{1/2}+T_{i,\mathrm{ur}}+2\sqrt{T_{i,\mathrm{ur}}\ln N^{1/2}}}}\Bigr\},
    \end{align*}
    and using Eq.~\eqref{eq:lem2:1} and~\eqref{eq:lem2:2}, we obtain
    \begin{equation*}
        P\left(\|\boldsymbol{\varepsilon}_{\mathrm{ur}}^i\|<\rho_{\mathrm{ur}}\right)>1-\dfrac{1}{\sqrt{N}}, \forall i,
        \qquad
        P\left(\|\boldsymbol{\varepsilon}_{\mathrm{co}}\|<\rho_{\mathrm{co}}\right)>1-\dfrac{1}{\sqrt{N}}.
    \end{equation*}
    
    Finally, we complete the proof as follows:
    \begin{align*}
        \bigl[\mathcal{L}_{\mathcal{D}}(\boldsymbol{\theta}^k)\bigr]_{k=1}^m
        & \le \mathbb{E}_{\boldsymbol{\varepsilon}_{\mathrm{co}}}\Bigl[\mathbb{E}_{\boldsymbol{\varepsilon}_{\mathrm{ur}}^k} \bigl[\mathcal{L}_{\mathcal{S}} (\boldsymbol{\theta}_{\mathrm{co}} +\boldsymbol{\varepsilon}_{\mathrm{co}},\boldsymbol{\theta}_{\mathrm{ur}}^k+\boldsymbol{\varepsilon}_{\mathrm{ur}}^k)\bigr]+f^k\left(\|\boldsymbol{\theta}^k\|_2^2\right)\Bigr]_{k=1}^m \\
        & \le \max_{\|\boldsymbol{\varepsilon}_{\mathrm{co}}\|<\rho_{\mathrm{co}}}\Biggl[\max_{\|\boldsymbol{\varepsilon}_{\mathrm{ur}}^k\|<\rho_{\mathrm{ur}}} \mathcal{L}_{\mathcal{S}} (\boldsymbol{\theta}_{\mathrm{co}} +\boldsymbol{\varepsilon}_{\mathrm{co}},\boldsymbol{\theta}_{\mathrm{ur}}^k+\boldsymbol{\varepsilon}_{\mathrm{ur}}^k)+\frac{2}{\sqrt{N}} \\
        & \qquad \qquad \qquad -\frac{1}{N}+f^k\left(\|\boldsymbol{\theta}^k\|_2^2\right)\Biggr]_{k=1}^m.
    \end{align*}
    To reach the final conclusion, we redefine
    \begin{equation*}
        f^k\left(\|\boldsymbol{\theta}^k\|_2^2\right) = \frac{2}{\sqrt{N}}-\frac{1}{N}+f^k\left(\|\boldsymbol{\theta}^k\|_2^2\right).
    \end{equation*}
    Here, we note that the last inequality follows from the decompositions (i.e., ball chunking)
    \begin{align*}
        & \mathbb{E}_{\boldsymbol{\varepsilon}_{\mathrm{co}}}\Bigl[\mathbb{E}_{\boldsymbol{\varepsilon}_{\mathrm{ur}}^k} \bigl[ \mathcal{L}_{\mathcal{S}} (\boldsymbol{\theta}_{\mathrm{co}}+\boldsymbol{\varepsilon}_{\mathrm{co}}, \boldsymbol{\theta}_{\mathrm{ur}}^k+\boldsymbol{\varepsilon}_{\mathrm{ur}}^k)\bigr]\Bigr]_{k=1}^m \\
        \le & \int_{B_{\mathrm{co}}}\Bigl[\int_{B_{\mathrm{ur}}^k} \mathcal{L}_{\mathcal{S}} (\boldsymbol{\theta}_{\mathrm{co}}+\boldsymbol{\varepsilon}_{\mathrm{co}}, \boldsymbol{\theta}_{\mathrm{ur}}^k+\boldsymbol{\varepsilon}_{\mathrm{ur}}^k)d\boldsymbol{\varepsilon}_{\mathrm{ur}}^k\Bigr]_{k=1}^m d\boldsymbol{\varepsilon}_{\mathrm{co}} \\
        & + \Bigl(1-\frac{1}{\sqrt{N}}\Bigr)\frac{1}{\sqrt{N}}+\frac{1}{\sqrt{N}} \\
        \le & \underbrace{\max_{\|\boldsymbol{\varepsilon}_{\mathrm{co}}\|<\rho_{\mathrm{co}}} \Bigl[\max_{\|\boldsymbol{\varepsilon}_{\mathrm{ur}}^i\|<\rho_{\mathrm{ur}}} \mathcal{L}_{\mathcal{S}} (\boldsymbol{\theta}_{\mathrm{co}} +\boldsymbol{\varepsilon}_{\mathrm{co}}, \boldsymbol{\theta}_{\mathrm{ur}}^k +\boldsymbol{\varepsilon}_{\mathrm{ur}}^k) \Bigr]_{k=1}^m}_{\displaystyle\text{worst-case empirical loss (SAM)}} +\frac{2}{\sqrt{N}}-\frac{1}{N},
    \end{align*}
    where $B_{\mathrm{co}}=\{\boldsymbol{\varepsilon}_{\mathrm{co}}:\|\boldsymbol{\varepsilon}_{\mathrm{co}}\|\le\rho_{\mathrm{co}}\}$, $B_{\mathrm{co}}^c$ is its complement, $B_{\mathrm{ur}}^k=\{\boldsymbol{\varepsilon}_{\mathrm{ur}}^k:\|\boldsymbol{\varepsilon}_{\mathrm{ur}}^k\|\le\rho_{\mathrm{ur}}\}$.
\end{proof}

\section{Visualization of Loss Landscape} \label{sec:loss_landscape}

Figure~\ref{fig:loss_landscape_appendix} presents the contour map visualization of the post-convergence landscapes for models trained with \method{} and with FedNCF.

Table \ref{tab:average_loss} summarizes the average loss values across different perturbation magnitudes.
The results indicate that \method{} converges to a flatter region of the loss surface.
On the FilmTrust dataset, for example, the average loss remains around 0.22 at a perturbation magnitude of 0.1, but increases drastically to $2 \times 10^{19}$ at 1, highlighting the sharpness of the FedNCF landscape.

\begin{table}[]
    \centering
    \caption{Average loss values under different perturbation magnitudes for FedNCF and \method{} across all datasets.}
    \resizebox{\linewidth}{!}{
        \begin{tabular}{cccccc}
            \toprule
            Methods & Magnitude & FilmTrust & Lastfm-2K & Amazon-Video & QB-article \\ \midrule
            \multirow{6}{*}{FedNCF} & 10 & $\sim$ 2e19 & 0.6504 & 0.6211 & 0.4268 \\
             & 1 & $\sim$ 2e19 & 0.7506 & 0.6128 & 0.4179 \\
             & 1e-1 & 0.2200 & 0.7662 & 0.6210 & 0.4037 \\
             & 1e-2 & 0.2207 & 0.7678 & 0.6219 & 0.4047 \\
             & 1e-3 & 0.2445 & 0.7679 & 0.6220 & 0.4051 \\
             & 1e-4 & 0.2206 & 0.7679 & 0.6220 & 0.4051 \\ \midrule
            \multirow{6}{*}{\method{}} & 10 & 0.2671 & 0.6593 & 0.5602 & 0.3317 \\
             & 1 & 0.4229 & 0.6397 & 0.5517 & 0.3221 \\
             & 1e-1 & 0.4150 & 0.6387 & 0.5506 & 0.3179 \\
             & 1e-2 & 0.4136 & 0.6386 & 0.5505 & 0.3179 \\
             & 1e-3 & 0.4133 & 0.6386 & 0.5505 & 0.3179 \\
             & 1e-4 & 0.4133 & 0.6386 & 0.5505 & 0.3179 \\
            \bottomrule
        \end{tabular}
    }
    \label{tab:average_loss}
\end{table}